\documentclass[twoside,11pt]{article}

\usepackage{dmlr2e}
\usepackage{lastpage}

\usepackage{booktabs}
\usepackage{longtable}
\usepackage{array}
\usepackage{adjustbox}
\usepackage{bm}
\usepackage{mathtools}
\usepackage{subcaption}
\let\origsubref\subref
\renewcommand{\subref}[1]{(\origsubref{#1})}
\usepackage[ruled,vlined]{algorithm2e}
\usepackage{mdframed}
\usepackage{listings}

\usepackage[capitalise]{cleveref}
\crefname{figure}{Fig.}{Figs.}
\Crefname{figure}{Fig.}{Figs.}
\crefname{table}{Table}{Tables}
\Crefname{table}{Table}{Tables}
\crefname{section}{Sec.}{Secs.}
\Crefname{section}{Sec.}{Secs.}
\crefname{algorithm}{Algorithm}{Algorithms}
\Crefname{algorithm}{Algorithm}{Algorithms}

\DeclareMathOperator*{\argmax}{argmax}

\SetKwComment{PythonStyleComment}{\# }{}

\SetCommentSty{mycommfont}

\newmdtheoremenv[
  linewidth=1pt,
  linecolor=black,
  backgroundcolor=white,
]{dfn}{Definition}

\newcommand{\eg}{e.g., }
\newcommand{\ie}{i.e., }

\dmlrheading{1}{2026}{1-\pageref{LastPage}}{01/26}{08/26}{XX-XXXX}{Takeda, Lin, Nozawa, Ng, Torii, and Matsui}
\def\openreview{\url{https://openreview.net/forum?id=xW9cXbbMJJ}}
\ShortHeadings{CIRCLED: Multi-turn CIR Dataset}{Takeda et al.}
\firstpageno{1}

\begin{document}

\title{CIRCLED: A Multi-turn CIR Dataset with Consistent Dialogues across Domains}

\author{\name Tomohisa Takeda$^1$ \email t\_takeda@hal.t.u-tokyo.ac.jp \\
       \name Yu-Chieh Lin$^2$ \email yuchieh.lin@kioxia.com \\
       \name Yuji Nozawa$^2$ \email yuji1.nozawa@kioxia.com \\
       \name Youyang Ng$^2$ \email youyang.ng@kioxia.com \\
       \name Osamu Torii$^2$ \email osamu.torii@kioxia.com \\
       \name Yusuke Matsui$^1$ \email matsui@hal.t.u-tokyo.ac.jp \\
       \addr $^1$Graduate School of Information Science and Technology, The University of Tokyo, Tokyo, Japan\\
       $^2$Kioxia Corporation, Tokyo, Japan}

\editor{Hugo Jair Escalante}

\maketitle

\begin{abstract}

Existing Multi-Turn Composed Image Retrieval (MTCIR) datasets lack dialogue-history consistency and are restricted to the fashion domain. 
To address these limitations, we construct CIRCLED by extending FashionIQ, CIRR, and CIRCO. 
In CIRCLED, the query at each turn progressively approaches the target image.
Data are generated via a CIReVL-based retrieval pipeline and curated with multiple filters on retrieval success, turn length, consistency, and information redundancy to ensure quality. In total, we collect 22,608 multi-turn sessions across nine subsets, substantially exceeding Multi-turn FashionIQ (11,505 sessions) in both scale and generality. We further apply multiple baseline methods and quantitatively assess retrieval accuracy on CIRCLED. Our work provides a practical, high-quality benchmark to facilitate future research on multi-turn CIR.
The dataset is publicly available at \url{https://huggingface.co/datasets/tk1441/CIRCLED}, and the code at \url{https://github.com/mti-lab/circled}.

\end{abstract}


\section{Introduction}
\label{sec:intro}

\begin{figure}[t]
  \centering
   \includegraphics[width=\linewidth]{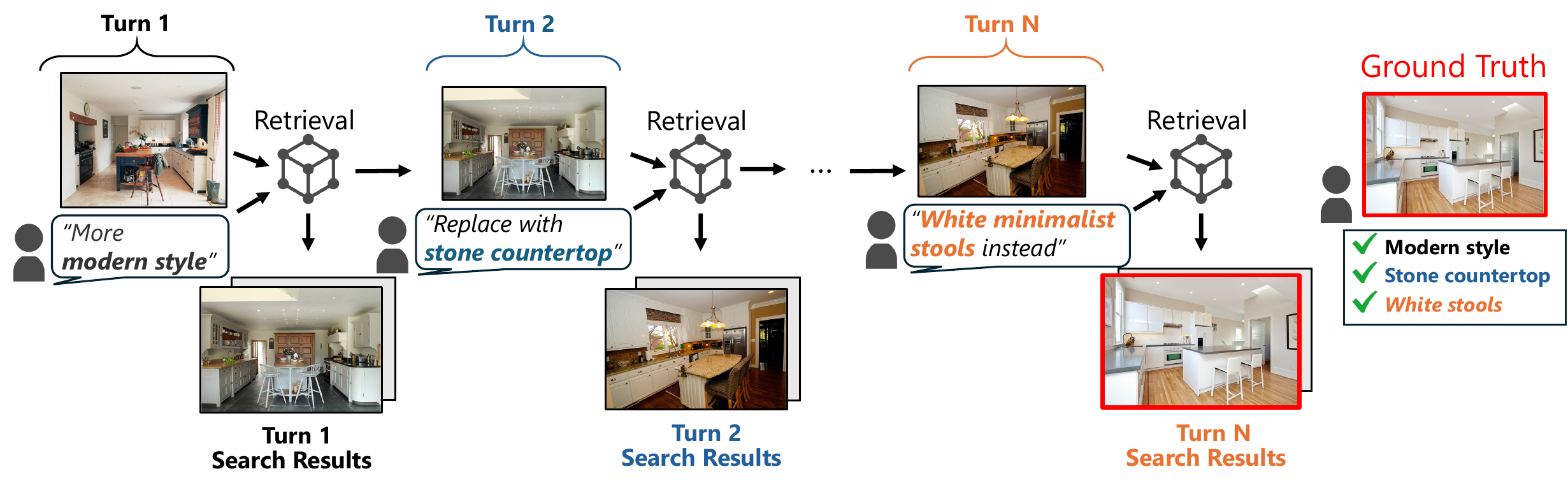}

   \caption{Example of multi-turn CIR search.
When users cannot find the image they want, they can search again in the next turn based on the existing search results to find the image they want.
The input for each turn is the same as the conventional CIR query, which is an image and text.}
   \label{fig:examples_multiturn_cir}
\end{figure}

Image retrieval aims to find images that match a user's intent in large corpora.
Traditional text-to-image~\citep{visualsparta, blip2} and content-based~\citep{deepfashion, spatial-bag-of-features} methods search with either a text query or an image, but they struggle when intent is ambiguous or nuanced.

Multimodal retrieval addresses this by combining text and images in a shared embedding space with pretrained vision-language models such as CLIP~\citep{CLIP} and BLIP~\citep{BLIP}.
Composed Image Retrieval (CIR) further refines user's intent by conditioning on a reference image plus a text modification, preserving visual structure while flexibly steering the search~\citep{fashion-cir, magiclens, llm-cir, bi-cir}.

Nevertheless, most CIR systems use a single-turn setting and give limited consideration to multi-turn usage, \ie interaction that progressively clarifies the user's intent.
This single-turn setup forces users to state their intent in one shot.
It performs poorly when user's intent is vague or when users refine requirements while inspecting results.

Against this background, multi-turn CIR~\citep{multiturn-fashioniq, fashionntm, mai}, which follows the flow illustrated in \cref{fig:examples_multiturn_cir}, has recently attracted attention.
In multi-turn CIR, retrieval is performed based on the history of $L$ image-text pairs $\{(I_1, T_1), \ldots, (I_L, T_L)\}$ used in previous turns, thereby more faithfully capturing the user's evolving intent.

\begin{figure}[t]
\centering

\begin{subfigure}{\linewidth}
    \centering
    \includegraphics[width=\linewidth]{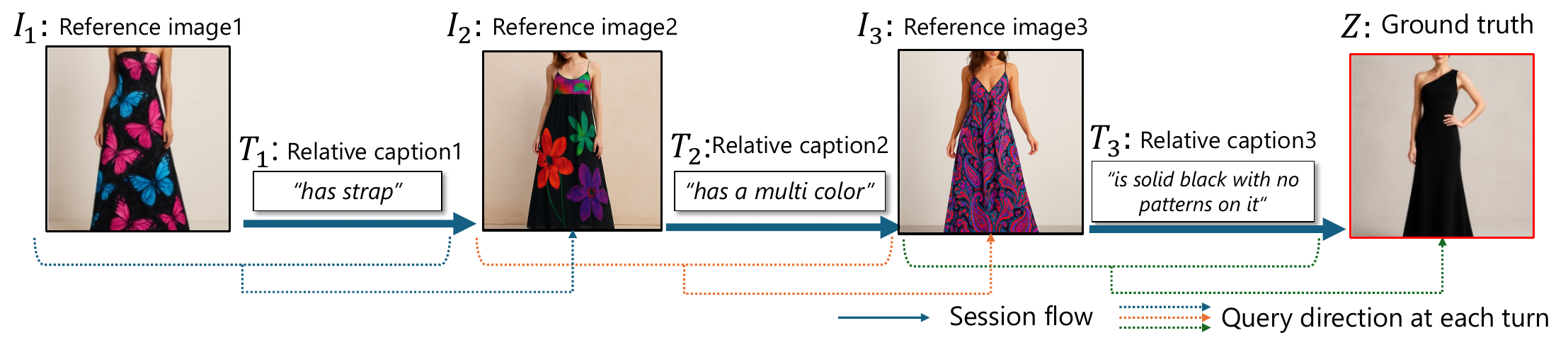}
    \caption{An example from an existing multi-turn CIR dataset}
    \label{examples_of_multiturn_fashioniq}
\end{subfigure}
\begin{subfigure}{\linewidth}
    \centering
    \includegraphics[width=\linewidth]{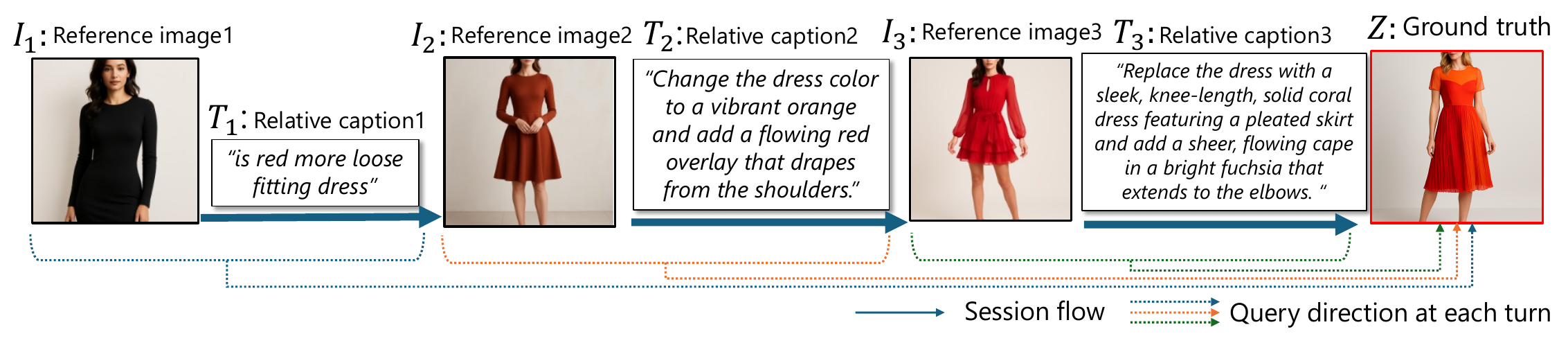}
    \caption{An example from our multi-turn CIR dataset CIRCLED}
    \label{examples_of_our_datasets}
\end{subfigure}
\caption{Comparison of multi-turn CIR datasets. Solid arrows indicate the progression of a session; dotted arrows indicate which previous inputs a given turn refers to. \subref{examples_of_multiturn_fashioniq} In the existing Multi-turn FashionIQ, feedback is primarily oriented toward the next turn's image and does not necessarily yield a consistent progression toward the ground truth\protect\footnotemark. \subref{examples_of_our_datasets} In our dataset, the query at each turn consistently points toward the ground truth, yielding a structure that progressively approaches it.}
\label{datasts_examples}
\end{figure}

A key issue in existing multi-turn CIR is inconsistent dialogue history.
At present, the only publicly available dataset is Multi-turn FashionIQ~\citep{multiturn-fashioniq}.
It is built by simply concatenating single-turn queries.
As shown in \cref{examples_of_multiturn_fashioniq}, it does not provide a structure in which each turn progressively approaches the ground truth (GT).
For example, even when the GT is a black dress, the dataset may include unrelated text such as ``has a strap'' or ``has a multi color,'' and assign later-turn reference images that are colorful dresses far from the intended black dress.
These intermediate-turn mismatches undermine realistic iterative search because later images and captions diverge from the target.
Moreover, prior work has focused on fashion, leaving multi-turn CIR in general domains largely unexplored.

We introduce CIRCLED, a multi-turn dataset with consistent, unidirectional progress toward the GT (\cref{examples_of_our_datasets}).
We design the query sequence $\{(I_1, T_1), \ldots, (I_L, T_L)\}$ so each turn steadily approaches the GT, mirroring real search behavior.
Each turn $(I_i, T_i)$ remains aligned with the GT, preserving coherence at both turn and session levels.
We construct CIRCLED by extending FashionIQ~\citep{fashioniq}, CIRR~\citep{cirr}, and CIRCO~\citep{circo}, broadening coverage beyond fashion to general domains.
CIRCLED contains 22,608 sessions spanning 2--6 turns and 202,845 images (\cref{tab:turn_distribution_and_category}), roughly twice that of Multi-turn FashionIQ, and includes 1,078 sessions with 5--6 turns, providing longer interactions.

We conduct several baseline experiments on CIRCLED to analyze its characteristics.
Furthermore, we assess performance using three metrics to complement aspects that previous multi-turn CIR settings could not evaluate.

The main contributions of this work are as follows:
\begin{itemize}
    \item CIRCLED: a multi-turn CIR dataset with consistency and progressive information gain, missing in prior work.
    \item Broader coverage beyond fashion by extending FashionIQ, CIRR, and CIRCO.
    \item Three metrics (Hits@10, Final Recall@10, AUC) enabling analysis of turn-wise success and early reachability, as well as final accuracy.
\end{itemize}

The dataset is publicly available at \url{https://huggingface.co/datasets/tk1441/CIRCLED}, and the construction and evaluation code at \url{https://github.com/mti-lab/circled} (see \cref{sec:dataset_documentation} for documentation).

\footnotetext[1]{For licensing reasons, the images shown in the figure are generated images that are visually similar to images in FashionIQ.}

\section{Related Work}
\label{sec:related_work}

\begin{table}[t]
\centering
\adjustbox{max width=\linewidth}{
\begin{tabular}{@{}lrrrrrrrl@{}} \toprule
& \multicolumn{6}{c}{\textbf{\#Sessions by Turn Length}} & & \\ \cmidrule(lr){2-7}
\textbf{Dataset} & \textbf{2} & \textbf{3} & \textbf{4} & \textbf{5} & \textbf{6} & \textbf{Total} & \textbf{\#Images}& \textbf{Categories} \\
\midrule
Multi-turn FashionIQ{\scriptsize~\citep{multiturn-fashioniq}} & 8,649 & 2,216 & 640 & -- & --  & 11,505 & 16,712 & {\small Dress, Shirt, Toptee} \\
CIRCLED (Ours)  & 14,216 & 5,424 & 1,890 & 772 & 306 & 22,608 & 202,845 & {\small Dress, Shirt, Toptee, General} \\
\bottomrule
\end{tabular}
}
\caption{Comparison of the number of sessions (=queries) by dialogue length $L$ and covered categories in CIRCLED and Multi-turn FashionIQ. $L$ denotes the number of turns per session; larger $L$ generally indicates higher difficulty. CIRCLED covers $ 2 \le L \le 6$ and includes long sessions with 5--6 turns, whereas Multi-turn FashionIQ is mostly $L\le 3$ (maximum $L=4$) and contains no 5--6 turn sessions. \textit{Total} is the total number of sessions, \textit{\#Images} is the number of database images, and \textit{Categories} lists the covered categories.}
\label{tab:turn_distribution_and_category}
\end{table}

\subsection{Composed Image Retrieval}

Composed Image Retrieval (CIR)~\citep{pic2word, magiclens, la-cir, bi-cir} retrieves a ground-truth image using a reference image together with a text modification.
By integrating visual and textual information, CIR enables flexible retrieval refinement in line with the user's intent.

Traditional CIR methods train on triplets (reference image, text, GT image), which demands large-scale datasets and costly training~\citep{mapping-network, magiclens}.
To reduce dependence on labeled data, recent works leverage Vision-Language Models (VLMs) and Large Language Models (LLMs) for zero-shot CIR~\citep{pic2word, circo, llm-cir, shimada}.
Most existing CIR methods use a single turn with one reference image and one text.
Iterative search is simulated by repeatedly issuing single-turn queries.
However, an explicit multi-turn framework that references past retrieval history and continuously tracks user intent has received relatively little attention.

\subsection{Multi-turn Composed Image Retrieval}

Multi-turn CIR is the task of retrieving the final target (ground truth) using the history of image-text pairs
$\{(I_1, T_1), \ldots, (I_L, T_L)\}$ as the query.

\citet{multiturn-fashioniq} pioneered multi-turn CIR by extending FashionIQ~\citep{fashioniq} and releasing Multi-turn FashionIQ (11,505 sessions, 3 categories), the first dataset for this setting.
It is built by concatenating single-turn pairs: each turn $(I_i, T_i)$ is designed as a query for the next image $I_{i+1}$.
However, there is no guarantee that intermediate images $\{I_2, \ldots, I_{L-1}\}$ progressively approach the ground truth, making intermediate turn evaluation infeasible.
Consequently, existing methods only report final-turn retrieval metrics.

\citet{fashionntm} construct MT Shoes (4,097 sessions, 10 categories) by concatenating single-turn transactions from the Shoes dataset~\citep{shoes} and propose a cascaded memory network to retain historical information.
\citet{mai} build FashionMT, a significantly larger dataset with 247,911 sessions, 95 categories, and 1,067,688 images, using LLM-based modification generation.
FashionMT introduces ``retrospective'' settings where users may refer back to attributes from previous turns (e.g., ``keep the color from turn 2'') or roll back to earlier images.
Despite its scale and retrospective design, FashionMT evaluates only the final turn and lacks explicit consistency constraints across turns, as intermediate images are not guaranteed to monotonically improve toward the ground truth.
Moreover, to the best of our knowledge and as of submission, the constructed multi-turn session data for MT Shoes\footnote{\url{https://sites.google.com/eng.ucsd.edu/fashionntm}} and FashionMT\footnote{\url{https://openreview.net/forum?id=gXyWbl71n1}} have not been released for download (only papers and supplementary material are available). In contrast, Multi-turn FashionIQ~\citep{multiturn-fashioniq} is publicly released.\footnote{\url{https://github.com/yfyuan01/MultiturnFashionRetrieval}} All three datasets remain limited to the fashion domain.

Beyond CIR datasets, interactive and conversational image retrieval has also been studied as a \emph{method}. ChatIR~\citep{ChatIR} and PlugIR~\citep{PlugIR} retrieve images through multi-turn natural-language dialogue with an LLM. UniCVR~\citep{unicvr} unifies single-turn, multi-turn, and video composed retrieval in a zero-shot framework. These works propose retrieval methods rather than a consistent multi-turn benchmark, which is our focus.

In contrast, our CIRCLED dataset ensures consistent progression toward the ground truth, within a bounded margin, via $\varepsilon$-consistency ($r_{l+1} \leq r_l + \varepsilon$; defined in \cref{subsec:dataset_characteristics}), enabling evaluation at any intermediate turn.
This allows us to introduce turn-wise metrics (Hits@10, AUC) that measure retrieval quality throughout the entire multi-turn session, not just the final turn.
Furthermore, CIRCLED is publicly available and covers both fashion and general domains (CIRR, CIRCO), offering broader applicability.

\Cref{tab:dataset_comparison} summarizes how CIRCLED compares with existing multi-turn CIR datasets. It is the only one that is simultaneously consistent, cross-domain, and publicly available.

\begin{table}[t]
\centering
\caption{Comparison of multi-turn CIR datasets. ``Consistent'' indicates a consistent rank progression toward the ground truth, within a bounded margin ($\varepsilon$-consistency). ``Public'' indicates that the constructed multi-turn session data is available for download as of submission. Only CIRCLED is simultaneously consistent, cross-domain, and public.}
\label{tab:dataset_comparison}
\adjustbox{max width=\linewidth}{
\begin{tabular}{@{}lcccc@{}} \toprule
Dataset & \#Sessions & Domain & Consistent & Public \\ \midrule
Multi-turn FashionIQ~\citep{multiturn-fashioniq} & 11{,}505  & Fashion           & $\times$     & \checkmark \\
MT Shoes~\citep{fashionntm}                      & 4{,}097   & Fashion           & $\times$     & $\times$ \\
FashionMT~\citep{mai}                            & 247{,}911 & Fashion           & $\times$     & $\times$ \\
\textbf{CIRCLED (ours)}                          & 22{,}608  & Fashion + General & \checkmark   & \checkmark \\ \bottomrule
\end{tabular}
}
\end{table}

In a related but distinct setting, \citet{chatsearch} propose ChatSearch for general conversational image retrieval with free-form multi-modal dialogues.
In contrast, our work extends CIR to multi-turn while preserving the structured query format of (reference image, text modification) pairs.

\section{Proposed Dataset: CIRCLED}
\label{sec:our_dataset}

We address two limitations of existing multi-turn CIR datasets: the lack of history consistency and the restriction to the fashion domain.
Concretely, we extend three single-turn CIR datasets (FashionIQ~\citep{fashioniq}, CIRR~\citep{cirr}, and CIRCO~\citep{circo}) to construct a new dataset that features consistent multi-turn dialogues across multiple domains.
CIRR~\citep{cirr} is a general-domain CIR dataset of real-life images requiring fine-grained distinctions, and CIRCO~\citep{circo} is an open-vocabulary CIR dataset built on COCO images with multiple ground truths per query. We process each source dataset \emph{independently} into multi-turn sessions rather than merging them, and we evaluate each subset separately. The general domain (CIRR and CIRCO) accounts for $34.9\%$ of all sessions and the fashion domain (FashionIQ) for $65.1\%$.

Summary statistics are shown in \cref{tab:turn_distribution_and_category}.
Compared to the existing dataset~\citep{multiturn-fashioniq}, CIRCLED is roughly twice as large in the number of queries and contains about twelve times more database images (largely reflecting the much larger general-domain search spaces of CIRR and CIRCO rather than more annotation per query), and it adds a General category beyond the three fashion categories. It also includes longer dialogues with 5--6 turns.

\subsection{Overview}
We describe the components of the dataset.
Let $\mathcal{X}=\{X_n\}_{n=1}^N$ be a database of $N$ images to be searched.
We define a \textit{session} as an $L$-turn query sequence $\{(I_1, T_1), (I_2, T_2), \dots, (I_L, T_L)\}$, where each $I_l$ is a reference image and each $T_l$ is a relative caption describing the desired change.
Each session is paired with a ground truth (GT) image $Z\in\mathcal{X}$.
Given a session, a retrieval algorithm $S$ must find $Z$ from $\mathcal{X}$.
The final performance of $S$ is computed by averaging results over many sessions.
We precompute and release these sessions so that future users can evaluate their multi-turn retrieval algorithms.

We form each session as follows.
We choose a GT image $Z \in \mathcal{X}$ from an existing single-turn CIR dataset.
We then construct a multi-turn query sequence $\{(I_1, T_1), (I_2, T_2), \dots, (I_L, T_L)\}$ that progressively approaches $Z$.
For Turn 1, we use the dataset's reference image $I_1$ and its relative caption $T_1$.
For later turns, we generate $T_2, T_3, \dots$ using an LLM and select $I_2, I_3, \dots$ from the retrieval results; details are given in \cref{sec:baseline retrieval algorithm}.

A well-formed multi-turn relative caption should (i) reference the current (selected) image, (ii) describe a concrete change toward the goal, (iii) add information not already provided, and (iv) improve the target's rank. The last two requirements correspond to the $\tau$-diversity and $\varepsilon$-consistency properties that we formalize in \cref{subsec:dataset_characteristics}. Turns~2 onward are generated to meet all four requirements. Turn~1 is intentionally different. At Turn~1 the user has not yet seen any candidates and states a vague initial intent, so we reuse the dataset's original single-turn relative caption $T_1$, a short comparative description. From Turn~2, the user has selected an image from the results and issues a concrete corrective instruction. This asymmetry mirrors the staged shift of intent before and after inspecting results. Keeping the original $T_1$ also makes Turn~1 a standard single-turn CIR query, which enables direct comparison with existing single-turn methods and grounds the dataset in the original human-annotated CIR data.

This session simulates a realistic search scenario in which the user refines the query while inspecting intermediate results.
Specifically, if the first-turn search with $(I_1, T_1)$ does not rank $Z$ sufficiently high, the user selects a desirable image from the current top results as $I_2$, and provides an additional description $T_2$ for $I_2$.
We rerun retrieval with $(I_1, T_1), (I_2, T_2)$, and this process is repeated until $Z$ appears sufficiently high in the ranking.

\subsection{Evaluation Protocol}

\begin{algorithm}[t]
\DontPrintSemicolon
\KwIn{retrieval algorithm $S$}
\KwOut{rank sequence $r_1, \dots, r_L \in \{1, \dots, N\}$}
\For{$l \in \{1, \dots, L\}$}{
$r_l \gets S(\{(I_1, T_1), (I_2, T_2), \dots, (I_l, T_l)\})$
}
\Return{$r_1, \dots, r_L$}
\caption{Single-session evaluation protocol. Apply $S$ to the cumulative history up to turn $l$ to obtain $r_l$. The dataset fixes the query sequence.}
\label{alg:evaluation_protocol}
\end{algorithm}

\Cref{alg:evaluation_protocol} outlines the evaluation pipeline for a single session.
Let $S$ be the retrieval algorithm under evaluation.
We evaluate retrieval performance turn by turn.
The first-turn search uses $\{(I_1, T_1)\}$, and the resulting rank of $Z$ is denoted $r_1\in\{1,\dots,N\}$:
\begin{equation}
r_1 = S(\{(I_1, T_1)\}) .
\end{equation}
At the second turn, we search using the accumulated information $\{(I_1, T_1), (I_2, T_2)\}$, yielding rank $r_2$.
Repeating this gives, at turn $l$,
\begin{equation}
r_l = S(\{(I_1, T_1), (I_2, T_2), \dots, (I_l, T_l)\}) . \label{ranking}
\end{equation}
The objective is to obtain a sufficiently small $r_l$ with as few turns (small $l$) as possible.

In this protocol, the dataset $\mathcal{X}$, each pair $(I_l, T_l)$, and the GT $Z$ are all fixed constants.
Importantly, the $I_l$ are predetermined and independent of the algorithm $S$, as in existing multi-turn CIR datasets.
Each session has one ground truth $Z$ (for CIRCO, a set of equivalent target images, of which the best-ranked one defines $r_l$). There is no per-turn ground truth. The intermediate $I_l$ are fixed query inputs, namely the reference images inspected during the simulated dialogue, and are not retrieval targets. Both the metrics (\cref{sec:evaluation metrics}) and $\varepsilon$-consistency are defined on the rank $r_l$ of this single $Z$ under the cumulative history, so they measure the same quantity. Fixing the $I_l$ makes the benchmark reproducible, since every algorithm $S$ is evaluated on the identical query sequence. This yields a controlled, non-interactive benchmark rather than a live user simulation. The $I_l$ were selected by the construction baseline, so the trajectory is fixed, but the same trajectory is used for every method, so the comparison remains fair.
If these intermediate choices are poor (\eg drifting from the GT or repeating information), the reported performance reflects dataset artifacts rather than the quality of $S$.
We guard against this with a strong baseline and the filtering procedures in ~\cref{sec:filtering process}.

The above describes evaluation for a single session.
We then apply this protocol to all sessions, compute the metrics defined in \cref{sec:evaluation metrics} (Hits@10, Recall@10, AUC) from the resulting rank sequences $\{r_l\}$, and report the mean over sessions as the final performance.

The protocol in \cref{alg:evaluation_protocol} generalizes many retrieval problems.
When $L=1$, it reduces to single-turn CIR.
When $T_1=T_2=\dots=\varnothing$, it becomes $L$-turn relevance feedback in image retrieval.
When $I_1=I_2=\dots=\varnothing$, it corresponds to $L$-turn chat-based image retrieval~\citep{ChatIR}
Thus, our protocol subsumes several existing settings as special cases, with multi-turn CIR as a particular instantiation.

\subsection{Dataset Characteristics}
\label{subsec:dataset_characteristics}
To realize natural multi-turn dialogues, we define two properties each session must satisfy: $\varepsilon$-consistency and $\tau$-diversity.
All sessions in CIRCLED satisfy these properties.

\begin{dfn}[\textbf{$\varepsilon$-consistency}]
A session is \emph{$\varepsilon$-consistent} w.r.t.\ a retrieval algorithm $S$ if its rank sequence $r_1, r_2, \dots$ satisfies
\begin{equation}
    r_{l+1} \le r_l + \varepsilon .
\end{equation}
\end{dfn}
Here, $\varepsilon$ is a small integer margin, and in CIRCLED we set $\varepsilon=30$.
With $\varepsilon$-consistency, we ensure that as turns progress, the GT (within a permitted margin) consistently moves toward higher ranks.
If a session violates this property, adding an off-target $(I_l, T_l)$ moves the aggregated query away from $Z$ in the embedding space, so repeating the query sequence can instead push the target image farther down the ranking.

Our dataset is constructed so that all sessions satisfy $\varepsilon$-consistency with respect to the baseline retrieval algorithm described in ~\cref{sec:baseline retrieval algorithm}.
This rank progression is not specific to the construction baseline. It also holds under other retrieval models and encoders (\cref{sec:construction_bias}).
By contrast, existing multi-turn CIR datasets do not satisfy this property; in some cases, the GT drifts downward as the dialogue proceeds, making it difficult to reliably evaluate retrieval algorithms. 

\begin{dfn}[\textbf{$\tau$-diversity}]
A session is \emph{$\tau$-diverse} w.r.t.\ a text encoder $E$ if its relative-caption sequence $T_1,\dots,T_L$ satisfies
\begin{equation}
    \max_{\,1 \le j < i \le L}\; \cos\!\big(E(T_i),\,E(T_j)\big) \;<\; \tau .
\end{equation}
\end{dfn}
Here, $E$ is a text encoder and $\cos(\cdot,\cdot)$ denotes cosine similarity.
With $\tau$-diversity, we ensure that each turn contributes novel information.
The threshold $\tau$ controls how much overlap is tolerated, and we use $\tau=0.8$ with $E$ a CLIP text encoder.
Smaller $\tau$ requires more novel information at each turn.
Without this constraint, the same or similar relative captions may be repeated across turns, adding no new information and preventing meaningful evaluation of multi-turn retrieval.

Each CIRCLED session exhibits a natural and consistent structure with these two properties: the GT rank steadily improves (allowing small fluctuations) as the dialogue progresses, and each turn adds new information.
In particular, for sessions with large $L$, $\tau$-diversity enforces continuous information addition, while $\varepsilon$-consistency ensures stepwise rank improvement, enabling clear visualization of performance differences in history aggregation for long dialogues.

\subsection{Dataset Bias Analysis}
\label{sec:bias_analysis}
Since the relative captions in CIRCLED are generated by an LLM, we investigate whether they exhibit different linguistic characteristics compared to existing single-turn CIR datasets (FashionIQ, CIRR, CIRCO), which were created by humans.
We follow prior caption analysis~\citep{fashioniq}.

\Cref{tab:linguistic_metrics} compares linguistic metrics.
CIRCLED's relative captions are longer but maintain comparable vocabulary diversity.
Specifically, CIRCLED's relative captions are significantly longer (average $|T_l| = 19.0$ words vs.\ 5.3--11.2 in existing datasets).
This difference stems from task design: single-turn CIR describes the overall difference between images at once, whereas multi-turn CIR requires explicit step-by-step modification instructions.
For example, single-turn may use concise descriptions like ``is darker and longer'' (4 words), while multi-turn requires specific instructions such as ``Replace the blue shirt with a black shirt featuring a centered graphic design'' (13 words).
The Type-Token Ratio measures vocabulary diversity as the ratio of unique words to total words~\citep{ttr}.
This metric is comparable between CIRCLED and existing datasets, suggesting that CIRCLED's vocabulary diversity remains competitive.
The part-of-speech ratios show the proportion of each word class in the captions.
FashionIQ exhibits a balanced distribution (Noun/Verb/Adj $\approx$ 29/28/28\%), whereas CIRCLED and CIRR/CIRCO are noun-dominant ($\approx$ 45/15/20\%).

FashionIQ and CIRR/CIRCO frequently use comparative expressions such as ``more,'' ``darker,'' and ``longer,'' while CIRCLED prominently features action verbs like ``add,'' ``replace,'' and ``change,'' as the word clouds of frequent terms we provide in \cref{sec:wordclouds} illustrate.
This reflects the difference in format: single-turn describes differences between images, whereas multi-turn provides step-by-step modification instructions.
Multi-turn FashionIQ exhibits a similar distribution to FashionIQ because it reuses the original FashionIQ captions without modification.

Differences across domains are also observed.
In the Fashion domain, both existing datasets and CIRCLED frequently use clothing terms such as ``sleeves,'' ``shirt,'' and ``dress,'' as well as color terms like ``black'' and ``blue.''
This is because tasks in the Fashion domain focus on fine-grained changes to clothing attributes (color, length, pattern, etc.).
In contrast, the General domain features more object and spatial terms such as ``dog'' and ``background.''
This reflects the prevalence of scene-level modifications in the General domain, such as adding/removing objects or changing backgrounds.

\begin{table}[t]
\centering
\caption{Linguistic metrics comparison.}
\label{tab:linguistic_metrics}
\begin{tabular}{@{}lcccc@{}} \toprule
    & \multicolumn{2}{c}{Fashion} & \multicolumn{2}{c}{General} \\
    \cmidrule(lr){2-3} \cmidrule(lr){4-5}
    Metric & FashionIQ & CIRCLED & CIRR/CIRCO & CIRCLED \\ \midrule
    Avg.\ Length of $T_l$ (words) & 5.3 & 19.0 & 11.2 & 19.0 \\
    Type-Token Ratio & 0.017 & 0.013 & 0.020 & 0.023 \\
    Noun Ratio (\%) & 29.3 & 45.4 & 46.4 & 45.3 \\
    Verb Ratio (\%) & 27.7 & 14.6 & 16.3 & 15.0 \\
    Adjective Ratio (\%) & 28.3 & 24.3 & 14.3 & 19.4 \\ \bottomrule
\end{tabular}
\end{table}

\section{Baseline Retrieval Algorithm} \label{sec:baseline retrieval algorithm}

\begin{figure}[t]
    \centering
    \begin{subfigure}{\linewidth}
        \centering
        \includegraphics[width=\linewidth]{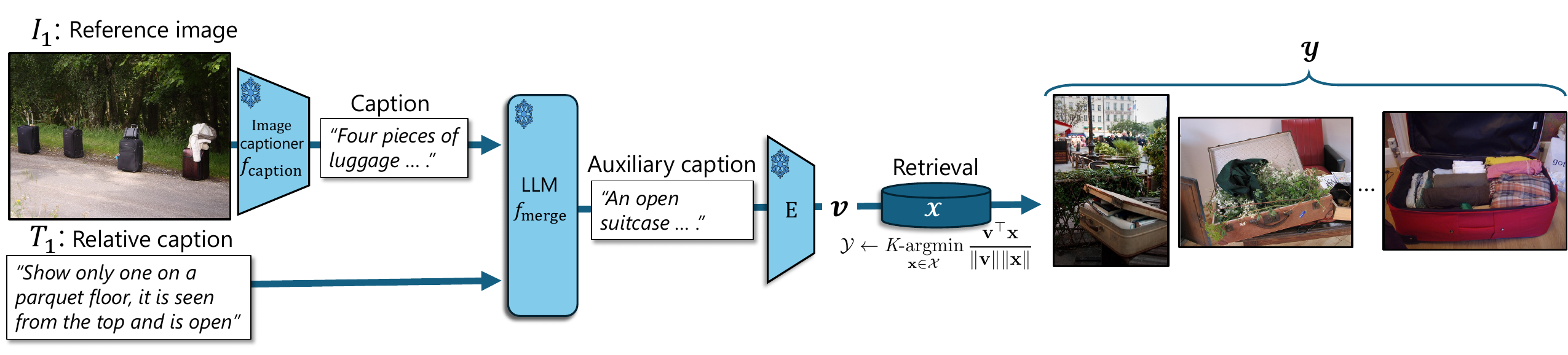}
        \caption{Search in Turn 1}
        \label{flow_turn0}
    \end{subfigure}
    \begin{subfigure}{\textwidth}
        \centering
        \includegraphics[width=\linewidth]{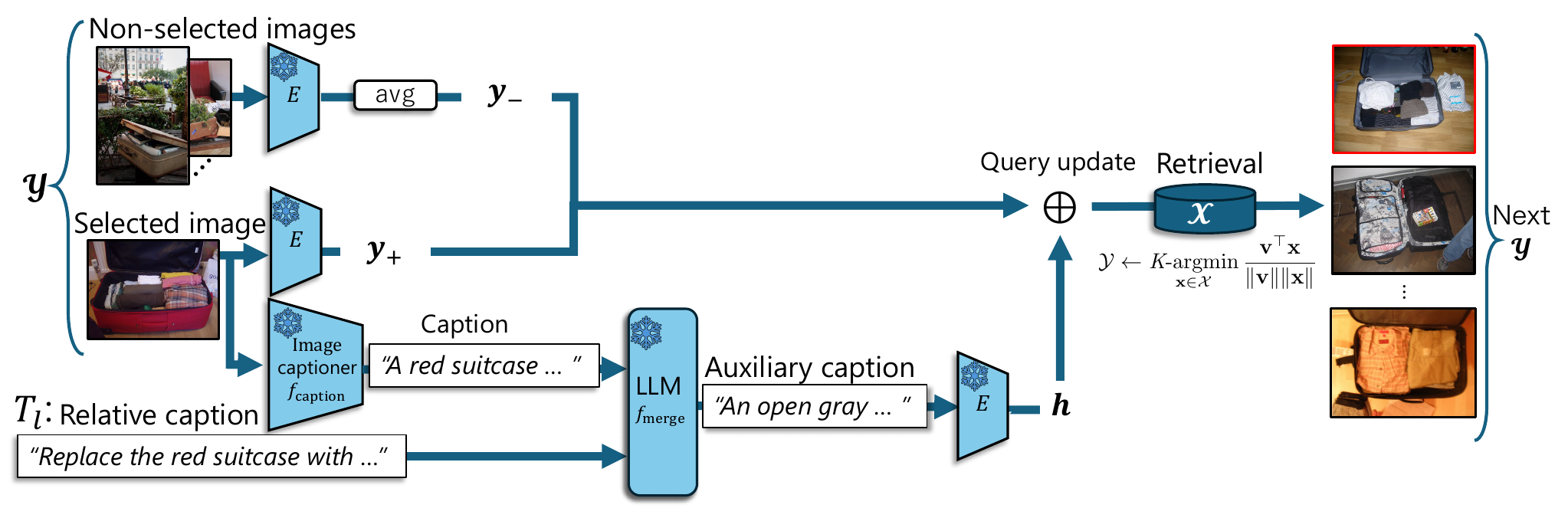}
        \caption{Search in Turn 2 and beyond}
        \label{flow_turn1}
    \end{subfigure}
    \caption{Multi-turn image retrieval pipeline.
\subref{flow_turn0} In Turn 1, we merge the caption generated from the reference image with the relative caption using an LLM and form a retrieval query following CIReVL~\citep{llm-cir}.
\subref{flow_turn1} From Turn 2 onward, we update the query by incorporating features of selected and non-selected images as well as textual information, and then perform image retrieval.}
    \label{flow_search_process}
\end{figure}

We describe the baseline retrieval algorithm $S_\mathrm{base}$ used to construct $(I_2, T_2), \dots$.
This baseline is a simple yet effective approach to multi-turn CIR and also decides ``which image to pick for the next turn'' when building the dataset.
Let $\mathbf{z} \in \mathbb{R}^D$ denote the $D$-dimensional feature of the GT image $Z$, and $\mathbf{x}_i \in \mathbb{R}^D$ the feature of image $X_i$.
For clarity, we redefine $\mathcal{X} = \{ \mathbf{x}_n \}_{n=1}^N$ as the set of image features.

\subsection{Algorithm Setup}
We extend CIReVL~\citep{llm-cir} to mimic human's multi-turn image search and to synthesize multi-turn CIR sessions.
CIReVL requires no training, is encoder-agnostic, and is independent of specific LLMs/VLMs, allowing flexible use with the latest models.
Briefly, CIReVL frames CIR as text-only retrieval: it captions the reference image and uses an LLM to fuse that caption with the relative caption into a single natural-language description, which is then embedded and matched against the image index.
In this paper, we implement it as follows:
\begin{itemize}
  \item Image/text embedding encoder: $E$ (BLIP~\citep{BLIP})
  \item Text merging: $f_{\mathrm{merge}}(\cdot,\cdot)$ (GPT-4o-mini~\citep{gpt-report})
  \item Image-to-caption generation: $f_{\mathrm{caption}}(\cdot)$ (GPT-4o-mini)
  \item Difference generator: $f_{\mathrm{diff}}(\cdot,\cdot,\cdot)$ (GPT-4o-mini)
\end{itemize}
We further ensure data quality with the filtering procedures described in ~\cref{sec:filtering process}.

\subsection{Retrieval Pipeline}
We start retrieval at Turn 1 and repeat it for up to six turns.
This six-turn cap is empirical rather than fundamental. Under the construction baseline, the ground truth enters the top-10 within six turns for the large majority of sessions, and only $1{,}078$ sessions ($4.8\%$) require five or six turns. Sessions that still fail after six turns are rare and are removed by the retrieval-success filter (\cref{sec:filtering process}).
Allowing more turns would mostly add long, hard-to-annotate sessions with diminishing returns and higher generation cost.
When the initial image $I_1$ is far from the target, this manifests as a high Turn-1 rank. Such a session simply needs more refinement turns to bring the GT into the top-10 (the retrieval-success condition). We retain it only if it converges within the six-turn budget, and otherwise filter it out.
The cap can be raised at construction time if longer interactions are desired.

\subsubsection{Turn 1 ($l=1$).}
As shown in \cref{flow_turn0}, Turn 1 follows the standard single-turn CIR setup:
given a reference image $I_1$ and a relative caption $T_1$, we generate a query and obtain the top-$K$ candidates $\mathcal{Y}$.
The steps are:
\begin{enumerate}
    \item Generate a caption from $I_1$ using $f_\mathrm{caption}$.
    \item Merge the generated caption with $T_1$ using $f_\mathrm{merge}$.
    \item Encode the merged sentence with the text encoder $E$ to obtain a feature vector $\mathbf{v} \in \mathbb{R}^D$.
    \item Use $\mathbf{v}$ to retrieve from $\mathcal{X}$ by cosine similarity and obtain the candidate set $\mathcal{Y} \subset \mathcal{X}$.
\end{enumerate}
If the GT $Z$ is not included in $\mathcal{Y}$, we consider the turn unsuccessful and proceed to the next turn.
For CIRCO, where multiple GT variants exist, we judge success if at least one of them appears in $\mathcal{Y}$.

\subsubsection{Turn 2 and beyond ($2 \le l$).}
From Turn 2 onward, we use the candidate set $\mathcal{Y}$ obtained in the previous turn and perform the following steps.

\paragraph{Selecting an image.}
We select from $\mathcal{Y}$ the image $y_+$ closest to the GT $Z$, excluding images chosen in earlier turns:
\begin{equation}
    \mathrm{sim}(Z, y) = \cos\!\left( E(f_\mathrm{caption}(Z)), \; E(f_\mathrm{caption}(y)) \right)
\end{equation}
\begin{equation}
    y_+ = \argmax_{y \in \mathcal{Y} \setminus \{I_1,\dots,I_{l-1}\}}\; \mathrm{sim}(Z, y) .
\end{equation}
Record this $y_+$ as the reference image $I_l$ for Turn $l$.
For CIRCO, we determine similarity using the caption of the ground-truth variant $Z$ most similar to the current query.

This step emulates user behavior: scanning the ranked list and picking the item closest to the GT.
Image-only selection often echoes the current top ranks and stalls progress, so we select with caption-based textual features.

\paragraph{Generating a relative caption.}
Next, we generate the relative caption $T_l$ from the selected image $y_+$, the GT image $Z$, and the history of past captions $H_{l-1}=(T_1,\dots,T_{l-1})$ by prompting a VLM:
\begin{equation}
    T_l \;=\; f_{\mathrm{diff}}(y_+,\, Z,\, H_{l-1}).
\end{equation}
We instruct the VLM to describe how $Z$ differs from $y_+$, avoid repeating $H_{l-1}$, and add new attributes or viewpoints.
For CIRCO, when multiple GT variants exist, we pass the variant $Z$ that is most similar to the current query.
This procedure mimics a user who issues iterative, corrective instructions based on the most recently selected image.

\paragraph{Updating the query.}
Finally, update the query using the following three components:
\begin{itemize}
    \item $\mathbf{y}_+$: feature of the selected image,
    \item $\mathbf{y}_-$: mean feature of the non-selected images $\big(\mathcal{Y} \setminus \{y_+\}\big)$,
    \item $\mathbf{h}$: feature of an auxiliary caption obtained by merging the selected image's caption with the new relative caption.
\end{itemize}
\begin{equation}
    \mathbf{h} = E\!\left(f_{\mathrm{merge}}\big(T_l,\, f_{\mathrm{caption}}(y_+)\big)\right)
\end{equation}
Update $\mathbf{v}$ as
\begin{equation}
\mathbf{v} \leftarrow \frac{\mathbf{v} + \alpha \mathbf{y}_{+} - \beta \mathbf{y}_{-} + \gamma \mathbf{h}}
{\left\lVert \mathbf{v} + \alpha \mathbf{y}_{+} - \beta \mathbf{y}_{-} + \gamma \mathbf{h} \right\rVert}.
\label{eq:query_update}
\end{equation}
where $\alpha, \beta, \gamma$ weight the terms.
Adding the selected image and auxiliary text while suppressing non-selected images aligns the update with user intent and curbs drift.

\subsection{Constructing Multi-turn Sessions}
By repeating this process for up to six turns, we extend an original single-turn CIR query
$(I_\mathrm{ref}, T_\mathrm{ref}) \rightarrow Z$
into an interactive sequence
$
(I_\mathrm{ref}, T_\mathrm{ref}) \rightarrow \dots \rightarrow (I_l, T_l) \rightarrow \dots \rightarrow Z.
$
Each $(I_l, T_l)$ is determined by the image-selection and relative-caption generation procedures described above.

\section{Dataset Construction}
Using the baseline $S_\mathrm{base}$ defined in the previous section, we extend existing single-turn CIR datasets and construct high-quality multi-turn dialogues.
This section details the filtering procedures and resulting statistics.

\subsection{Filtering Process} \label{sec:filtering process}
The multi-turn data generated with $S_\mathrm{base}$ may include failures (the target is never retrieved) or redundant relative captions that add no new information.
We therefore apply four filters to remove such cases.

\begin{table}
\centering
\adjustbox{max width=\linewidth}{
\begin{tabular}{@{}lrrrrrr@{}}
\toprule
{Subsets} & Init & $\triangle$ Succ & $\triangle$ Multi & $\triangle$ Rank & $\triangle$ Text & Final \\
\midrule
\textbf{fdt} & 10883 & 5829 & 1011 & 86 & 930 & 3027 \\
\textbf{fdv} & 3722 & 1468 & 491 & 17 & 386 & 1360 \\
\textbf{fst} & 11300 & 4656 & 2067 & 149 & 910 & 3518 \\
\textbf{fsv} & 3881 & 1216 & 740 & 54 & 371 & 1500 \\
\textbf{ftt} & 11544 & 4705 &  1698 & 123 & 1208 & 3810 \\
\textbf{ftv} & 3732 & 1011 & 707 & 32 & 476 & 1506 \\
\textbf{crt} & 28225 & 5847 & 13328 & 95 & 2081 & 6874 \\
\textbf{crv} & 4181 & 225 & 2801 & 4 & 192 & 959 \\
\textbf{cov} & 220 & 26 & 124 & 2 & 14 & 54 \\
\midrule
\textbf{Total}           & \textbf{77,688} & \textbf{24,983} & \textbf{22,967} & \textbf{562} & \textbf{6,568} & \textbf{22,608} \\
\bottomrule
\end{tabular}
}
\caption{Filtering pipeline: per-stage removals (stage-wise decrements) and final counts by subset.
Succ: retrieval-success; Multi: multi-turn; Rank: rank-margin; Text: text-redundancy.
Subsets: fdt/fdv/fst/fsv/ftt/ftv = FashionIQ Dress/Shirt/Toptee (train/val);
crt/crv = CIRR (train/val); cov = CIRCO (val).}
\label{filtering-removed-count}
\end{table}

\begin{table}[t]
\centering
\adjustbox{max width=\linewidth}{
\begin{tabular}{@{}lrrrrrrrr@{}}
\toprule
& \multicolumn{3}{c}{\textbf{Summary}} & \multicolumn{5}{c}{\textbf{\#Sessions by turn}} \\
\cmidrule(lr){2-4} \cmidrule(lr){5-9}
\textbf{Subset} & \textbf{\#Sessions (total)} & \textbf{\#Images} & \textbf{Avg. turns} & \textbf{2 turns} & \textbf{3 turns} & \textbf{4 turns} & \textbf{5 turns} & \textbf{6 turns} \\
\midrule
fashioniq\_dress\_train  & 3,027 & 10,886  & 2.74 & 1,583 &   920 &  323 & 144 &  57 \\
fashioniq\_dress\_val    & 1,360 &  3,653  & 2.73 &   745 &   358 &  163 &  64 &  30 \\
fashioniq\_shirt\_train  & 3,518 & 18,500  & 2.49 & 2,376 &   752 &  251 & 101 &  38 \\
fashioniq\_shirt\_val    & 1,500 &  6,182  & 2.52 &   990 &   333 &  110 &  48 &  19 \\
fashioniq\_toptee\_train & 3,810 & 15,742  & 2.62 & 2,253 & 1,002 &  359 & 126 &  70 \\
fashioniq\_toptee\_val   & 1,506 &  5,261  & 2.56 &   934 &   392 &  119 &  45 &  16 \\
cirr\_train              & 6,874 & 16,939  & 2.50 & 4,581 & 1,496 &  513 & 219 &  65 \\
cirr\_val                &   959 &  2,297  & 2.38 &   718 &   166 &   41 &  23 &  11 \\
circo\_val               &    54 & 123,385 & 2.61 &    36 &     5 &   11 &   2 &   0 \\
\midrule
\textbf{Total}           & \textbf{22,608} & \textbf{202,845} & \textbf{2.56} & \textbf{14,216} & \textbf{5,424} & \textbf{1,890} & \textbf{772} & \textbf{306} \\
\bottomrule
\end{tabular}
}
\caption{Post-filtering statistics: number of sessions, number of images, average turns, and distribution of sessions by turn count for each subset.
The  Subsets column follows the naming \{dataset\}\_\{category (dress, shirt, toptee)\}\_\{split (train or val)\}.}
\label{tab:dataset-summary}
\end{table}

\paragraph{1. Retrieval-success filter.}

\noindent We remove any query where the GT $Z$ never appears in the top-$K$ ($K=10$) candidates $\mathcal{Y}$ during turns 1--6.

\paragraph{2. Multi-turn filter.}

\noindent We exclude queries that already succeed at Turn 1 (i.e., the original single-turn query), keeping only dialogues that require at least two turns.

\paragraph{3. Rank-margin filter ($\varepsilon$-consistency).}

\noindent We remove any session that has a turn $l$ with $r_{l+1} > r_l + \varepsilon$.
We set $\varepsilon=30$.
This enforces $\varepsilon$-consistency and prevents large drops in the GT rank at intermediate turns.

\paragraph{4. Text-redundancy filter ($\tau$-diversity).}

\noindent We remove any session that has a turn $i$ where $\max_{1 \le j < i \le L}\cos\big(E(T_i),E(T_j)\big)\ge\tau$.
We set $\tau=0.8$ and use CLIP~\citep{CLIP} for $E$.
This enforces $\tau$-diversity and prevents caption redundancy across turns.

These four filters ensure both consistency in retrieval accuracy and the incremental addition of information in the relative captions. As summarized in \cref{filtering-removed-count}, four filters (success, multi-turn, rank-margin, and text-redundancy) cut 77,688 queries to 22,608 sessions.

\begin{figure}[t]
\centering
\begin{subfigure}{\columnwidth}
    \centering
    \includegraphics[width=\linewidth]{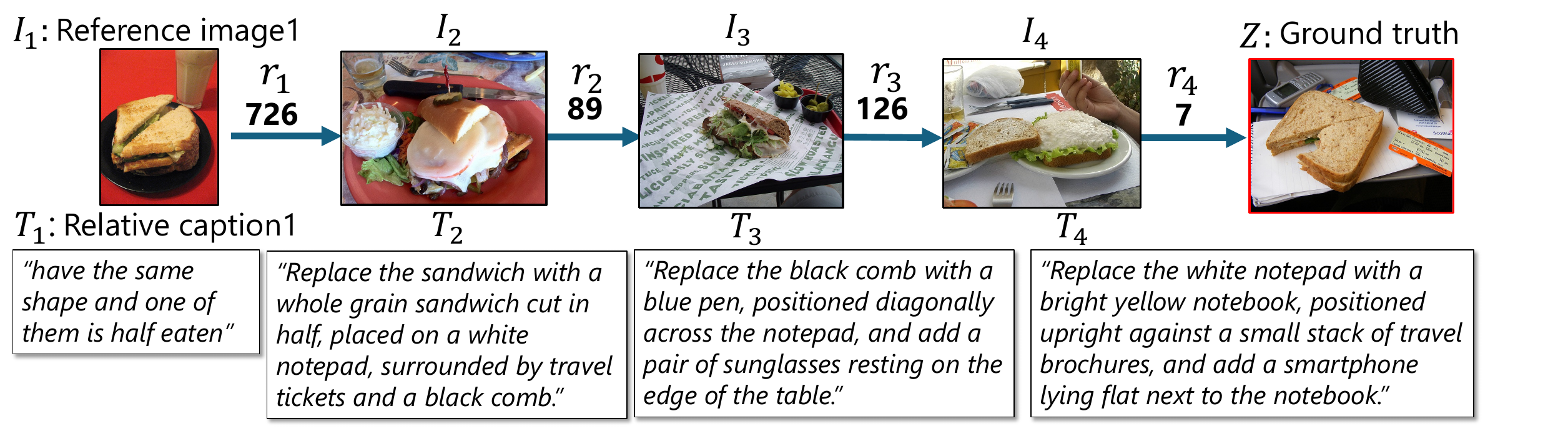}
    \caption{Rank-margin filter: mid-turn GT rank drop ($\varepsilon$-consistency violated).}
    \label{rank filter}
\end{subfigure}
\begin{subfigure}{\columnwidth}
    \centering
    \includegraphics[width=\linewidth]{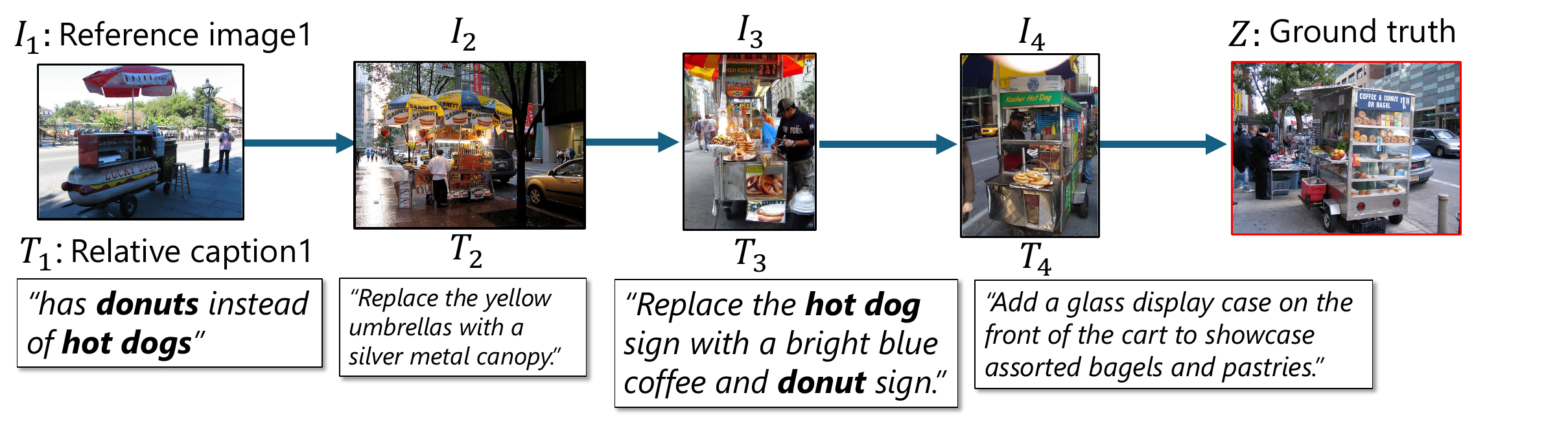}
    \caption{Text-redundancy filter: duplicate captions ($\tau$-diversity violated).}
    \label{relative caption filter}
\end{subfigure}
\caption{Examples rejected by our filtering. \subref{rank filter} Rank-margin filter ($\varepsilon=30$). \subref{relative caption filter} Text-redundancy filter ($\tau=0.8$).}
\label{filter examples}
\end{figure}

Examples are shown in \cref{filter examples}.
\Cref{rank filter} is removed because the GT rank significantly worsens middle dialogue; \cref{relative caption filter} is removed because the relative caption repeats earlier content.

\begin{figure}[t]
\centering
\begin{subfigure}{\columnwidth}
    \centering
    \includegraphics[width=\linewidth]{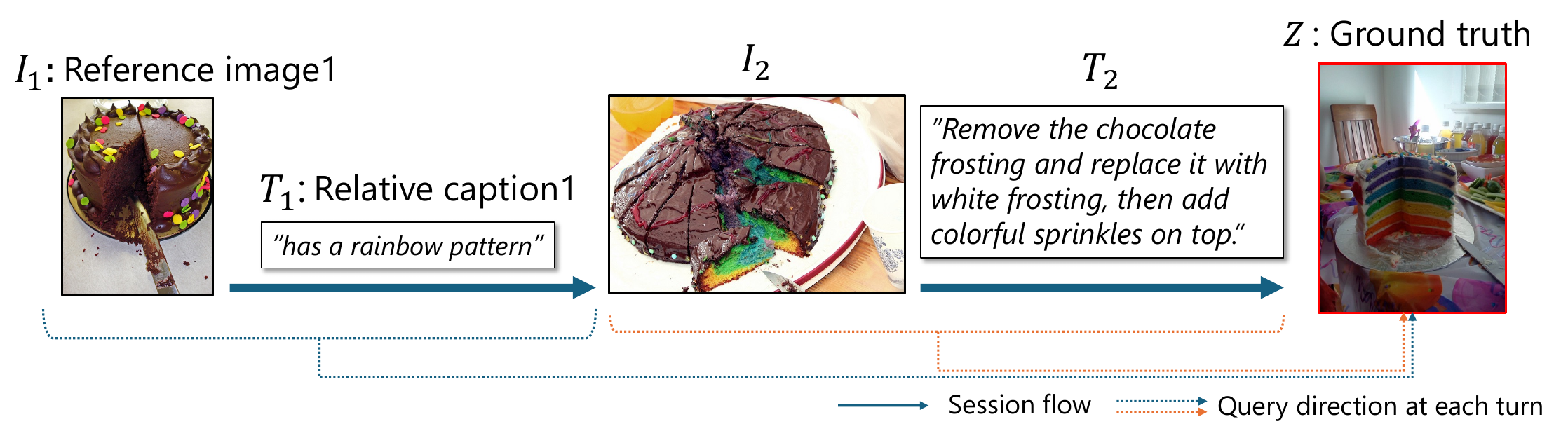}
    \caption{General domain (\texttt{circo\_val}): A 2-turn session progressively refining cake attributes}
    \label{fig:example_circo}
\end{subfigure}
\begin{subfigure}{\columnwidth}
    \centering
    \includegraphics[width=\linewidth]{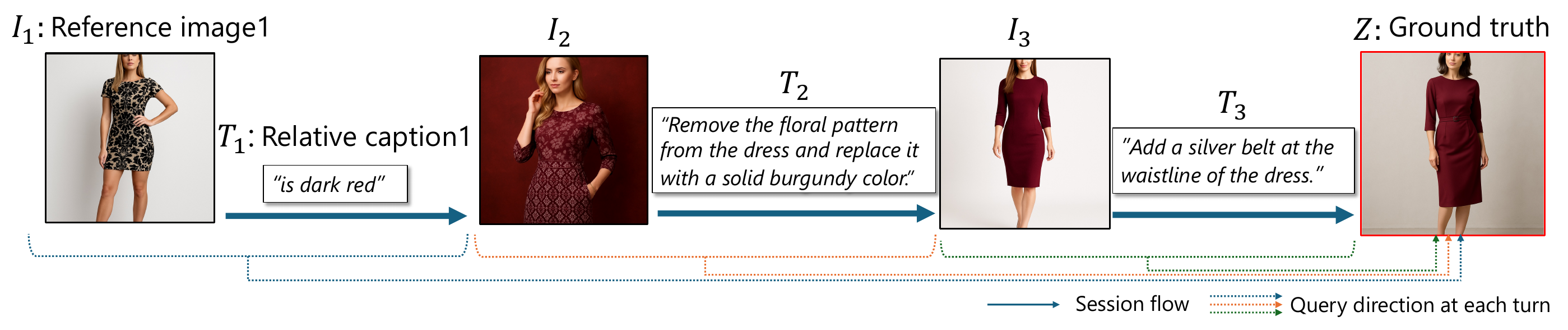}
    \caption{Fashion domain (\texttt{fashioniq\_dress\_val}): A 3-turn session progressively adding dress details}
    \label{fig:example_fashioniq}
\end{subfigure}
\caption{Examples of filtered CIRCLED sessions demonstrating gradual progression toward the ground truth.
\subref{fig:example_circo} A 2-turn session in the general domain (\texttt{circo\_val}).
\subref{fig:example_fashioniq} A 3-turn session in the fashion domain (\texttt{fashioniq\_dress\_val}).
Each turn combines visual (reference image $I_l$) and textual (relative caption $T_l$) information while satisfying $\varepsilon$-consistency and $\tau$-diversity\protect\footnotemark.}
\label{fig:successful_examples}
\end{figure}

In contrast, \cref{fig:successful_examples} shows examples of sessions that passed filtering.
These examples demonstrate how retrieval progressively narrows toward the ground truth by combining multimodal information (image and text) at each turn, in both the general domain (\texttt{circo\_val}) and the fashion domain (\texttt{fashioniq\_dress\_val}).

\subsection{Statistics}
We summarize the key statistics of the filtered dataset.
As shown in \cref{tab:dataset-summary}, sessions average 2.56 turns and span 2--6 turns.
CIRCLED adds a General category and outscales Multi-turn FashionIQ~\citep{multiturn-fashioniq} in queries, images, and category coverage (\cref{tab:turn_distribution_and_category}).

Regarding turn-length distribution (\cref{tab:dataset-summary}), approximately 63\% of sessions (14,216) have 2 turns, 24\% (5,424) have 3 turns, and only about 5\% (1,078) have 5--6 turns.
This distribution is consistent across domains: the proportion of 5--6 turn sessions is 5.1\% for Fashion and 4.1\% for General, showing no significant difference.

This short-session distribution results from our quality filters, particularly the text-redundancy filter ($\tau$-diversity).
As shown in \cref{filtering-removed-count}, the text-redundancy filter removed 6,568 sessions, approximately 12 times more than the 562 sessions removed by the rank-margin filter.
That is, most sessions terminate due to $\tau$-diversity violations (generating relative captions similar to previous turns) rather than $\varepsilon$-consistency violations (worsening GT rank).
To confirm this, we compared CIRCLED with sessions excluded by the rank-margin and text-redundancy filters (``w/o quality filter'' in \cref{tab:user_study}; these sessions pass the retrieval-success and multi-turn conditions).
As shown in \cref{tab:user_study}, the excluded sessions average 3.53 turns---about one turn longer than CIRCLED's 2.54---and contain a much higher proportion of 5--6 turn sessions.
As turns progress, fewer new attributes remain to describe differences between the GT and reference images, making redundant relative captions more likely.
Allowing longer sessions without filtering would increase turns with little information, degrading dataset quality.
Indeed, as detailed in \cref{sec:session_quality}, the excluded sessions score lower than CIRCLED on Coherence, Goal Progress, and Redundancy.
However, various biases in LLM-based evaluation have been reported, including position bias (sensitivity to presentation order)~\citep{llm-as-judge} and self-preference bias (favoring low-perplexity texts)~\citep{self-preference-bias}.

\footnotetext[2]{For licensing reasons, the images shown in the figure are generated images that are visually similar to images in FashionIQ.}

\subsection{Session Quality Evaluation with LLM-as-a-Judge}
\label{sec:session_quality}
To evaluate the quality of generated multi-turn dialogues, we conducted session-level evaluation using LLM-as-a-judge.
LLM-as-a-judge has been widely studied as an alternative to human evaluation~\citep{llm-judge-survey}, and with appropriate design, it has been reported to show high correlation with human evaluation~\citep{llm-as-judge,llm-as-judge-qa,judging-the-judges}.

Based on these findings, we adopted the following design:
(1) each session was evaluated independently to avoid ordering bias between comparison targets;
(2) we used GPT-5-mini as the evaluation model, different from GPT-4o-mini used for dataset generation, to avoid self-preference bias;
(3) each session was rated on five dimensions (naturalness, coherence, goal progress, low redundancy, overall) on a 1--5 scale.
We sampled approximately 60 sessions per subset from each source where available (1,711 sessions total).
Note that Multi-turn FashionIQ covers only Fashion subsets, and some sessions were excluded due to the Azure OpenAI API content filters.
We compared: (1) CIRCLED (proposed dataset), (2) w/o quality filter (sessions excluded by filters 3--4), (3) Multi-turn FashionIQ, and (4) Simple Concat (simple concatenation of single-turn datasets).

As shown in \cref{tab:user_study} (higher is better for all metrics), CIRCLED achieved the highest scores on Coherence, Goal Progress, Redundancy, and Overall.
The w/o quality filter set scored highest on Naturalness, but this is attributable to differences in turn-count distribution: it has an average of 3.53 turns, and longer sessions tend to receive higher naturalness scores.
When comparing sessions with the same number of turns, there was no significant difference in naturalness between CIRCLED and w/o quality filter.
Simple Concat scored lowest on all metrics, confirming that simple concatenation of single-turn datasets results in unnatural multi-turn dialogues.
We assess statistical significance with pairwise one-sided Mann-Whitney $U$ tests between CIRCLED and each comparison source (\cref{tab:significance}). CIRCLED significantly outperforms every comparison source on Overall, Coherence, Goal Progress, and Redundancy (all $p<0.001$). Effect sizes (rank-biserial $r$) are large against Simple Concat ($r=0.51$--$0.61$), small-to-medium against Multi-turn FashionIQ ($r=0.11$--$0.31$), and small against the w/o-quality-filter set ($r=0.08$--$0.19$). The only non-significant difference is Naturalness against the w/o-quality-filter set, which is consistent with its longer average turn count.
We further validate these LLM-judge ratings with a human study on the same 50 sessions (\cref{sec:human_eval}). Human annotators agree with the LLM judge at the session level (Spearman $\rho=0.65$ on Overall), reproduce the same source ranking exactly (Spearman $\rho=1.00$ over the four source means), and rate CIRCLED highest.

\begin{table}[t]
\centering
\caption{Session-level evaluation using LLM-as-a-judge (GPT-5-mini). Each metric is rated 1--5 (higher is better). The prompt is provided in \cref{sec:llm_judge_prompt}. ``w/o quality filter'' includes sessions excluded by filters 3--4 (rank-margin and text-redundancy) but passing filters 1--2.}
\label{tab:user_study}
\resizebox{\textwidth}{!}{
\begin{tabular}{@{}lccccccc@{}} \toprule
    Source & $N$ & Avg. Turns & Natural. & Coher. & Progress & Redund. & Overall \\ \midrule
    CIRCLED (Ours) & 463 & 2.54 & \underline{3.89}{\scriptsize$\pm$0.71} & \textbf{4.37}{\scriptsize$\pm$0.72} & \textbf{4.56}{\scriptsize$\pm$0.66} & \textbf{4.96}{\scriptsize$\pm$0.20} & \textbf{4.34}{\scriptsize$\pm$0.60} \\
    w/o quality filter & 463 & 3.53 & \textbf{4.00}{\scriptsize$\pm$0.61} & \underline{4.20}{\scriptsize$\pm$0.74} & \underline{4.28}{\scriptsize$\pm$0.79} & \underline{4.87}{\scriptsize$\pm$0.39} & \underline{4.17}{\scriptsize$\pm$0.63} \\
    Multi-turn FashionIQ & 336 & 2.30 & 3.51{\scriptsize$\pm$0.72} & 4.15{\scriptsize$\pm$0.83} & 4.08{\scriptsize$\pm$0.86} & 4.80{\scriptsize$\pm$0.53} & 3.98{\scriptsize$\pm$0.74} \\
    Simple Concat & 449 & 2.84 & 3.38{\scriptsize$\pm$0.81} & 3.47{\scriptsize$\pm$0.98} & 3.46{\scriptsize$\pm$0.99} & 4.67{\scriptsize$\pm$0.60} & 3.44{\scriptsize$\pm$0.82} \\ \bottomrule
\end{tabular}
}
\end{table}

\section{Experimental Results on CIRCLED}
\label{sec:experiment results}

In this section, we quantitatively and qualitatively evaluate retrieval performance on the proposed multi-turn CIR dataset.
We analyze how each turn affects accuracy and the relative utility of text and image information.

\subsection{Evaluation Metrics} \label{sec:evaluation metrics}
\noindent
For evaluation, let $\mathcal{Q}$ denote the set of evaluation sessions (queries).
We extend $r_l$ introduced in \cref{ranking} to each query $q\in\mathcal{Q}$, and denote by $r_l^{(q)}$ the rank of the (single) GT $Z$ for query $q$ at turn $l$, computed from the cumulative history up to turn $l$ (the same quantity used in $\varepsilon$-consistency).
When multiple ground truth targets exist (as in CIRCO), we take $r_l^{(q)}$ to be the best rank among them.
Let $L^{(q)}$ be the session length of query $q$, and let $L_{\max}=\max_{q\in\mathcal{Q}} L^{(q)}$ be the maximum length in the dataset.
We then define
\begin{equation}
    \mathrm{hit}_l^{(q)} = \mathbf{1}\!\big(r_l^{(q)} \le K\big), \quad K=10 ,
\end{equation}
where $\mathbf{1}(\cdot)$ denotes the indicator function: it equals $1$ if the GT is in the top-$K$ at turn $l$ and $0$ otherwise.
We use the following three metrics:
\begin{itemize}
    \item \textbf{Hits@10(turn $l$)}
\begin{equation}
    \mathrm{Hits@10}(l) \;=\; \frac{1}{|\mathcal{Q}|}
    \sum_{q\in\mathcal{Q}} \max_{1 \le l' \le l}
    \mathrm{hit}_{l'}^{(q)}
\end{equation}
    The fraction of queries that reach the top-10 at least once by turn $l$.
    This captures how quickly a method succeeds.
    This value is cumulative; for each query, if $l>L^q$ then $\mathrm{Hits@10}(l)=\mathrm{Hits@10}(L^q)$.
    \item \textbf{Final Recall@10}
    \begin{equation}
        \mathrm{Final\ Recall@10}
        = \frac{1}{|\mathcal{Q}|}
        \sum_{q\in\mathcal{Q}} \mathrm{hit}_{L^{(q)}}^{(q)} .
    \end{equation}
    Accuracy using all available information at the final turn; used to compare final performance with existing methods.
    \item \textbf{AUC over Hits@10}
\begin{equation}
    \mathrm{AUC} \;=\; \frac{1}{L_{\max}-1}
\sum_{l=1}^{L_{\max}-1}
\frac{\mathrm{Hits@10}(l) + \mathrm{Hits@10}(l+1)}{2}.
\end{equation}
    The trapezoidal integral of the Hits@10 curve.
    Larger values indicate reaching the top-10 in fewer turns, thus evaluating both final accuracy and convergence speed.
\end{itemize}

\subsection{Compared Methods}

Existing multi-turn CIR methods are difficult to use as fair baselines. MT Shoes~\citep{fashionntm} and FashionMT~\citep{mai} are trained models that release neither their code/pretrained weights nor their constructed datasets, so they can be neither run directly nor retrained, and Multi-turn FashionIQ's method~\citep{multiturn-fashioniq} is fashion-specific. A from-scratch re-implementation would not faithfully reproduce them and would make for an unreliable comparison. Instead, we compare reproducible, widely-used CIR methods with publicly available weights, and adapt each to the multi-turn setting in a uniform, training-free way.

\noindent\textbf{Multi-turn adaptation.} We do not modify any method. At turn $l$ each method produces a normalized feature $\bm{f}_l$. We aggregate the per-turn features over the cumulative history with one of three strategies (Latest, Average, and exponentially-decayed Weighted, detailed in \cref{sec:history_integration}), and retrieve with the aggregated query under \cref{alg:evaluation_protocol}. This aggregation is identical across methods, so differences reflect the underlying CIR method rather than bespoke multi-turn engineering.

\noindent\textbf{Training-free setting.} All baselines are used zero-shot. Text-only and Image-only are plain CLIP similarity. Pic2Word~\citep{pic2word} and MagicLens~\citep{magiclens} use their publicly released pretrained weights. CIReVL~\citep{llm-cir} composes CLIP retrieval with GPT-4o-mini~\citep{gpt-report} captioning and is training-free by design. None are fine-tuned on CIRCLED and the aggregation adds no learnable parameters, so CIRCLED serves as a \emph{zero-shot} evaluation benchmark. We use CLIP (ViT-L/14) as the encoder for all methods.

\noindent\textbf{Baselines:} Text-only (relative caption), Image-only (reference images), Pic2Word, CIReVL (GPT-4o-mini), and MagicLens.

\subsection{Feature Aggregation Strategies}
Let $\bm{f}_l$ (already normalized) be the feature at turn $l$.
We aggregate them as follows (exponential decay coefficient $\alpha=0.8$ in experiments):
\begin{flalign*}
& \textbf{Latest:} \quad \hat{\bm{f}} = \bm{f}_l & \\[-0.3em]
& \textbf{Average:} \quad \hat{\bm{f}} = \tfrac{1}{l}\textstyle\sum_{l'=1}^{l}\bm{f}_{l'} & \\[-0.3em]
& \textbf{Weighted:} \quad \hat{\bm{f}} = \tfrac{1}{C}\textstyle\sum_{l'=1}^{l}\alpha^{l-l'}\bm{f}_{l'}, \quad C=\textstyle\sum_{l'=1}^{l}\alpha^{l-l'} &
\end{flalign*}
We report results using the best aggregation strategy for each subset and method.
We provide a detailed analysis of aggregation strategies in \cref{sec:history_integration}.

\subsection{Quantitative Results}
\subsubsection*{Turn-wise Hits@10 Analysis}
\begin{figure}
    \centering
    \captionsetup{aboveskip=2pt, belowskip=0pt}
    \begin{subfigure}{0.32\linewidth}
        \centering
        \includegraphics[width=\linewidth]{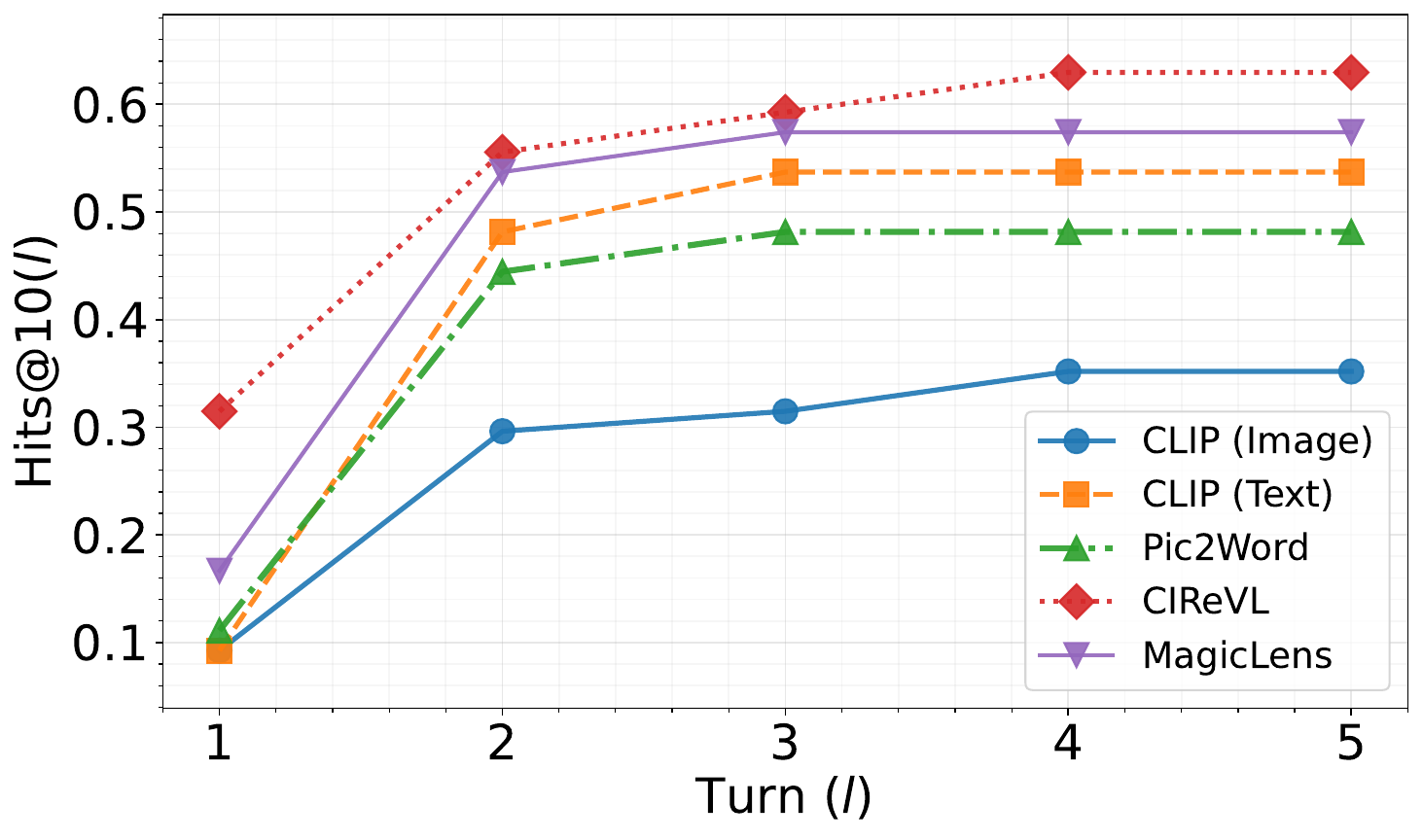}
        \caption{circo\_val}
        \label{circo_val_hits}
    \end{subfigure}%
    \begin{subfigure}{0.32\linewidth}
        \centering
        \includegraphics[width=\linewidth]{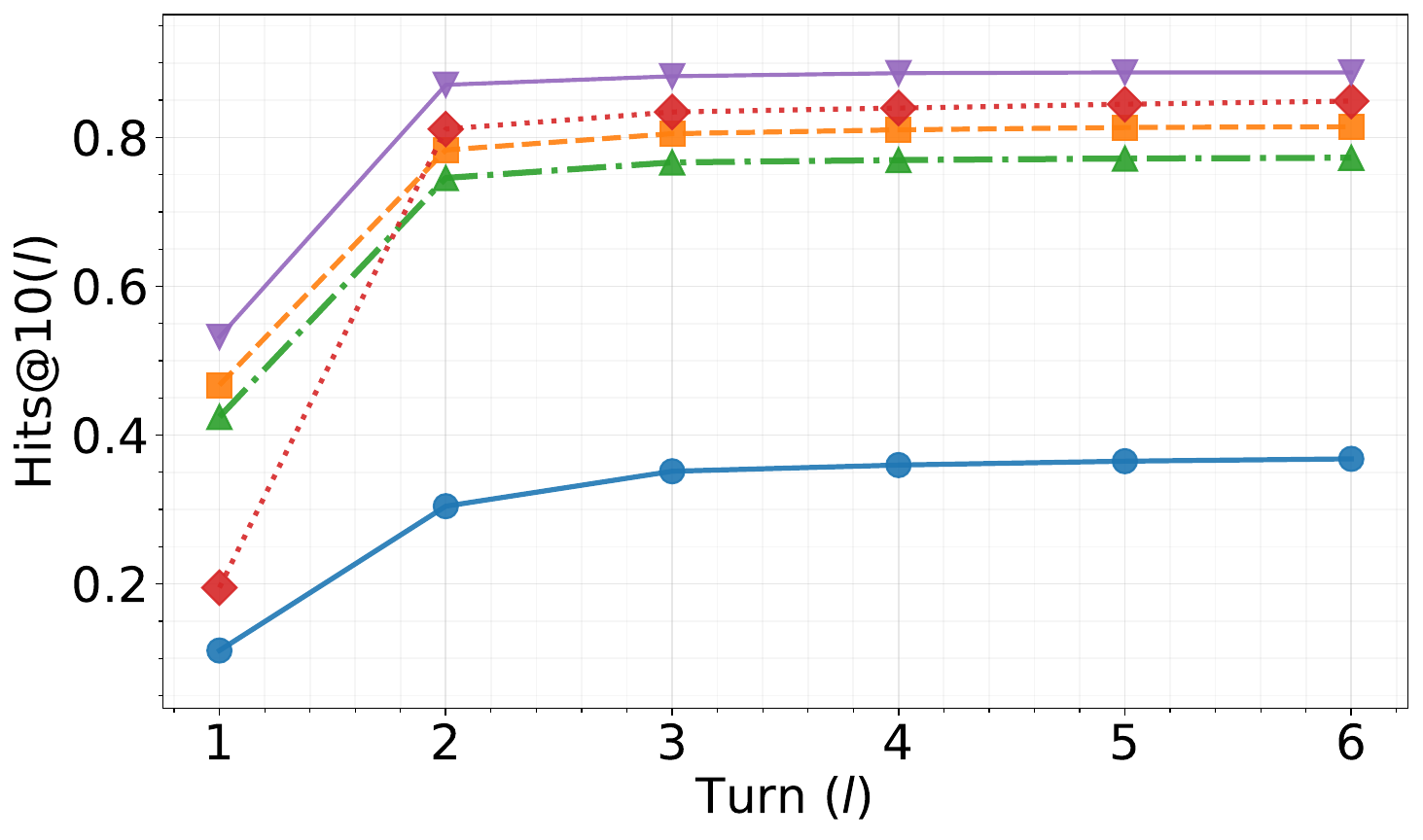}
        \caption{cirr\_val}
        \label{cirr_val_hits}
    \end{subfigure}%
    \begin{subfigure}{0.32\linewidth}
        \centering
        \includegraphics[width=\linewidth]{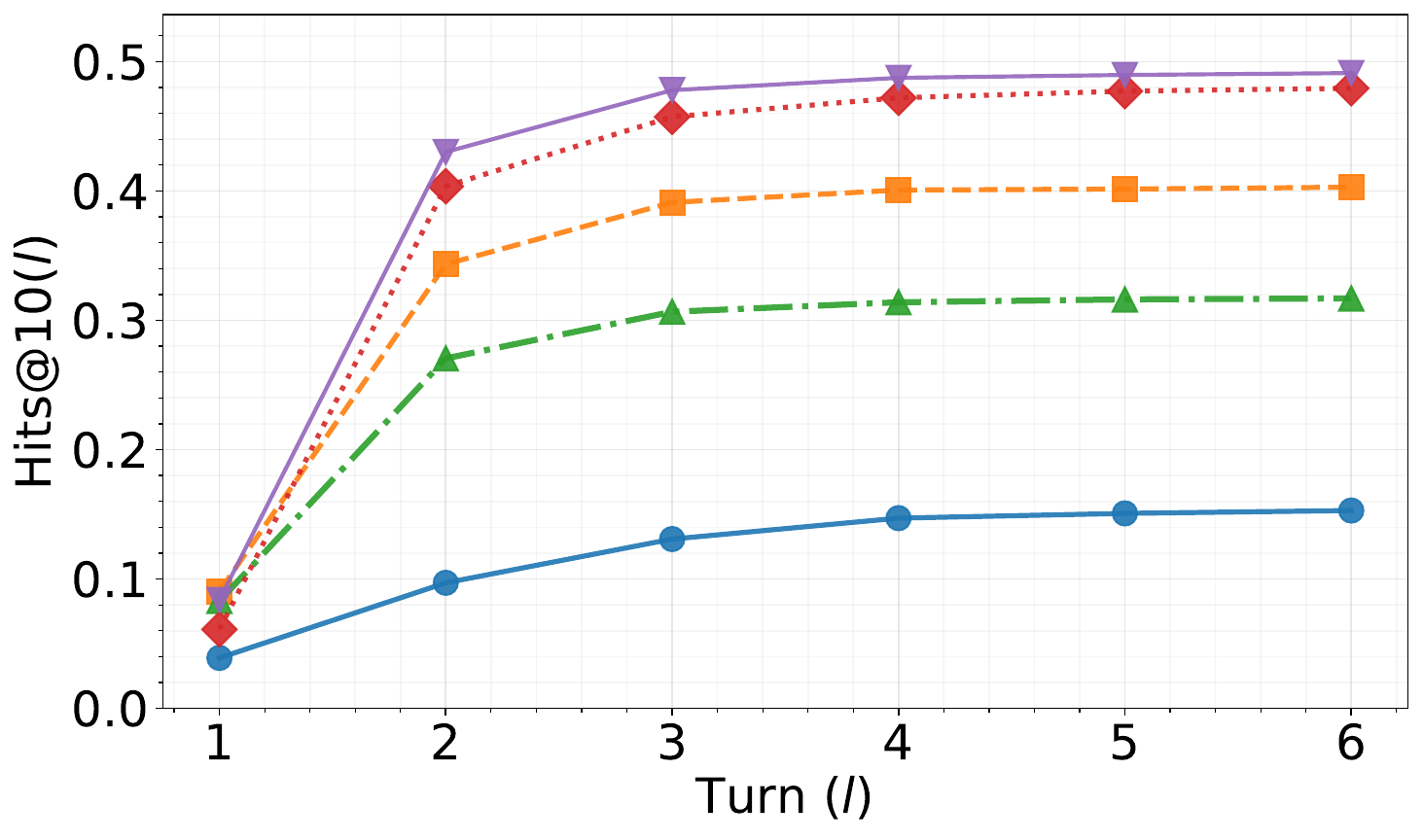}
        \caption{fashioniq\_dress\_val}
        \label{fashioniq_dress_val_hits}
    \end{subfigure}
    \caption{Comparison of Hits@10 by turn across subsets.}
    \label{htis_cir}
\end{figure}

Turn-wise Hits@10 results are shown in \cref{htis_cir}.
On \texttt{cirr\_val}, MagicLens exceeds 85\% but gains little with more turns.
On \texttt{circo\_val} and \texttt{fashioniq\_dress\_val}, accuracy climbs then plateaus near 60\% and 50\%.
Pic2Word trails Text-only, highlighting the value of text.

\subsubsection*{Final Recall@10 and AUC Analysis}
\begin{figure}
\centering
\begin{subfigure}{0.48\linewidth}
    \centering
    \includegraphics[width=\linewidth]{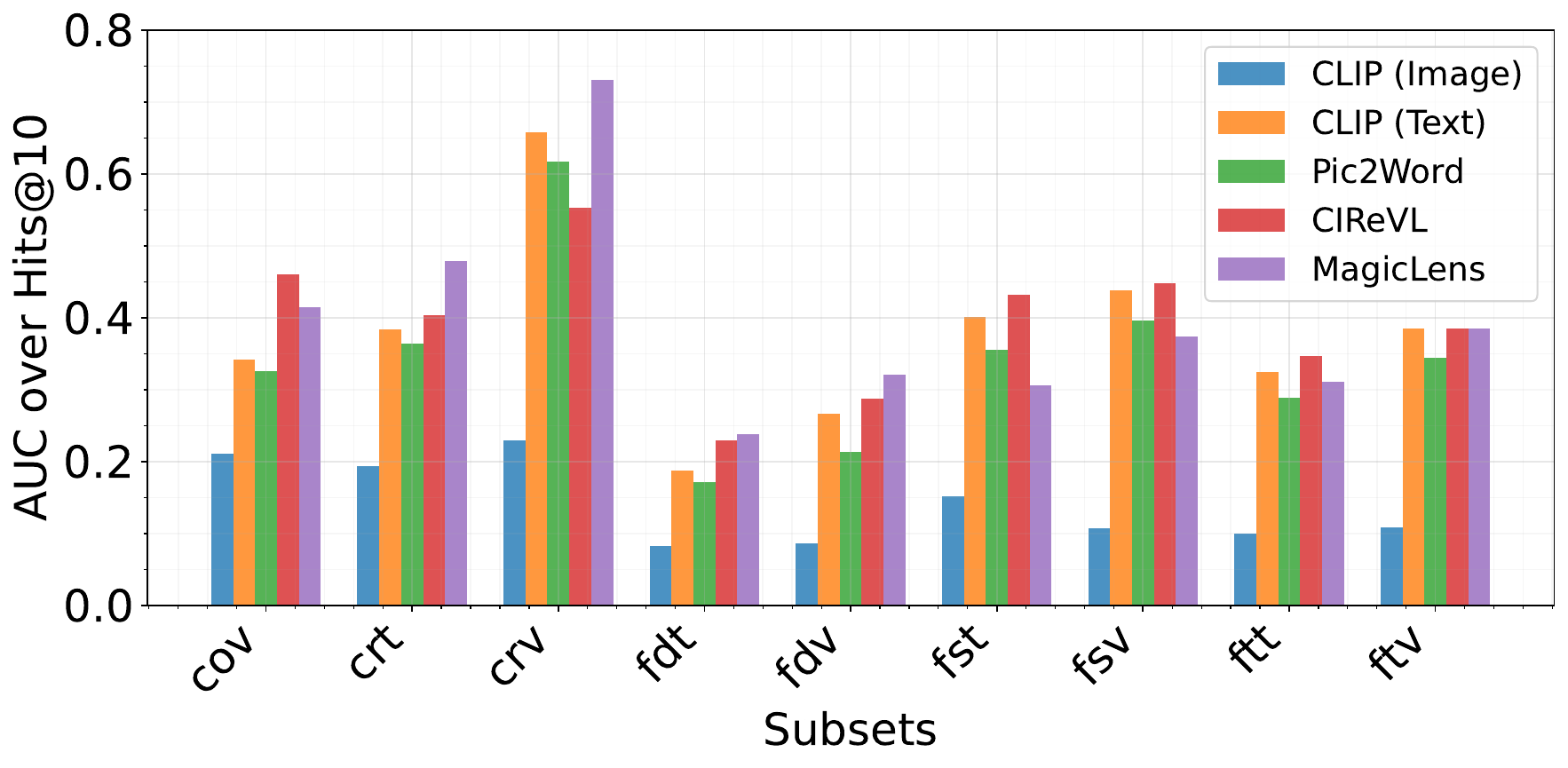}
    \caption{AUC over Hits@10}
    \label{auc}
\end{subfigure}
\hfill
\begin{subfigure}{0.48\linewidth}
    \centering
    \includegraphics[width=\linewidth]{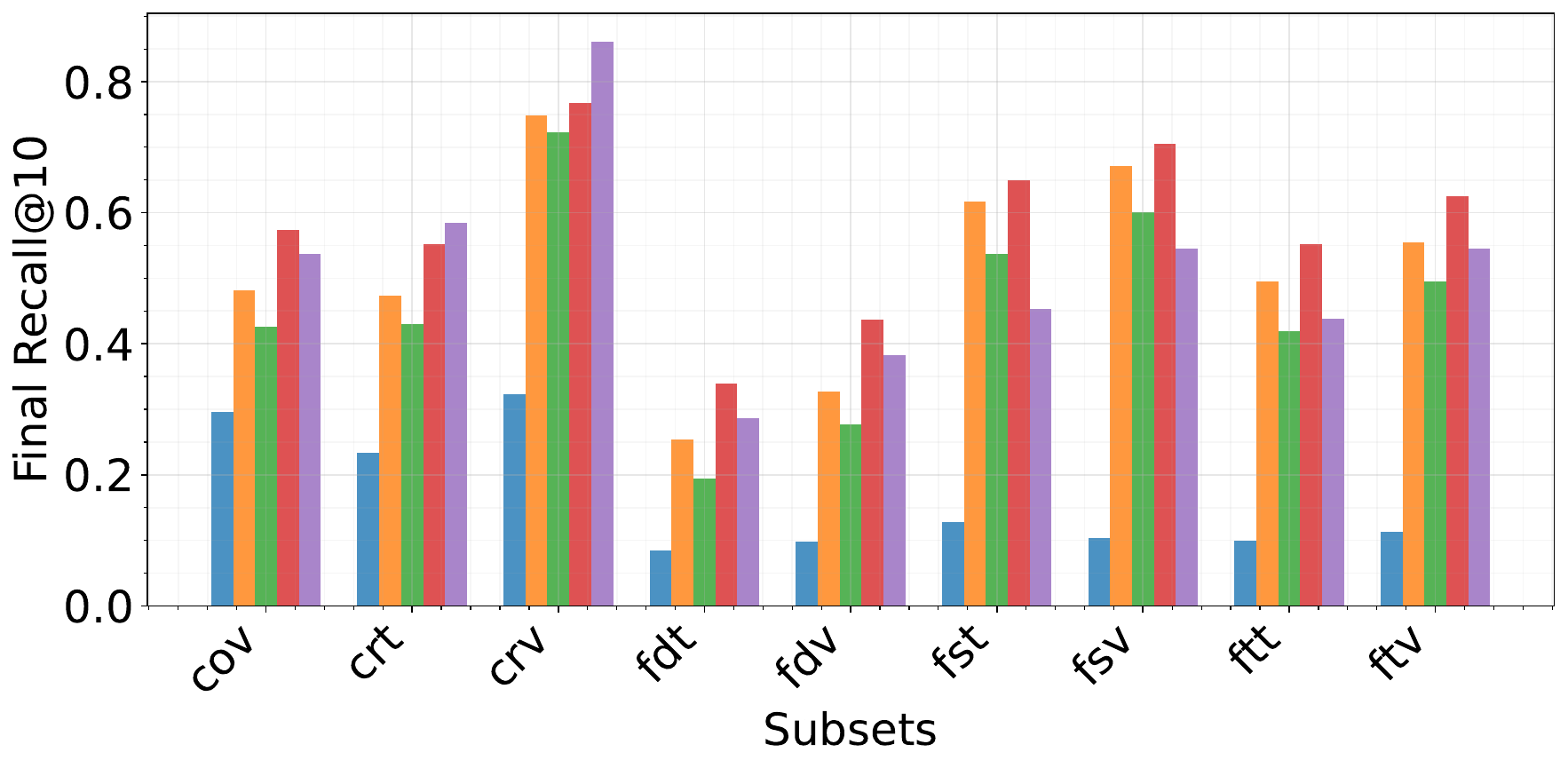}
    \caption{Final Recall@10}
    \label{final recall@10}
\end{subfigure}
\caption{Final Recall@10 and AUC over Hits@10 across baselines, by subset
(codes as in ~\cref{filtering-removed-count}).}
\label{final_recall_and_auc}
\end{figure}

Final Recall@10 and AUC results are shown in \cref{final_recall_and_auc}.
CIReVL, Text-only, and MagicLens perform well overall.
MagicLens leads on general-domain data, while CIReVL is stronger on fashion.
On \texttt{cirr\_val}, CIReVL beats Text-only on Hits@10 from Turn 2 onward. Its weak first turn lowers AUC, showing that AUC reflects performance across turns.
Thus, the preferred method depends on the metric and use case, so we report multiple metrics.

Image-only performs notably worse than Text-only across all subsets, confirming that relative captions carry more discriminative information than reference images alone in multi-turn CIR.
The domain-specific performance gap may stem from training data: MagicLens was trained on diverse web-crawled image pairs favoring general scenes, whereas CIReVL's text-based approach better captures fine-grained fashion attributes such as color, pattern, and silhouette.
\section{Conclusion}
\label{sec:conclusion}

We presented \textit{CIRCLED}, a multi-turn CIR dataset built by extending FashionIQ, CIRR, and CIRCO. It addresses two gaps in prior work: inconsistent dialogue histories and a fashion-only scope.
Each turn is designed to steadily approach the ground truth, ensuring coherence at both turn and session levels and enabling study beyond fashion.

We evaluated several baselines and found that the textual modality is the most informative and that the best-performing method depends on the domain.
We position CIRCLED as a \emph{controlled benchmark}. Its rank-progression structure holds across the retrieval methods we evaluate, not only the construction baseline. By construction, it is defined relative to an iterative-refinement retrieval algorithm, rather than a simulation of unconstrained human search.
CIRCLED provides a practical dataset and an evaluation framework for future research on multi-turn CIR.\footnote{The license information for the images used in this paper is provided in \cref{sec:image_licenses}.}

\impact{
This work provides a standardized benchmark for multi-turn composed image retrieval, contributing to the development of interactive visual search systems.
Potential applications include conversational product search in e-commerce and AI agents that progressively explore visual information to fulfill user requests.
By offering a consistent, cross-domain benchmark, CIRCLED highlights several design challenges it poses for future CIR methods. These methods must aggregate and model dialogue history across turns, remain robust beyond the fashion domain, and move from a fixed evaluation trajectory toward closed-loop interactive retrieval.
In terms of scalability, the baselines we study are training-free. At each turn they perform a single CLIP forward pass followed by nearest-neighbor search over the subset database, requiring no per-dataset fine-tuning, while construction uses a bounded, $O(L)$ number of LLM (GPT-4o-mini) calls per $L$-turn session. \Cref{tab:scalability} summarizes these costs and the main constraints.
Our dataset uses images from existing public datasets (FashionIQ, CIRR, CIRCO), thereby avoiding privacy concerns associated with new image collection.
However, since the relative captions are generated by an LLM (GPT-4o-mini), they may inherit linguistic biases present in the model, and the captions are English-only. In addition, the evaluation trajectory is fixed rather than interactive.
To mitigate the bias risk, we fully disclose the dataset construction process (prompts, filtering criteria, etc.), enabling the research community to verify and address potential biases.
}

\acks{
This work was not supported by any specific grant from funding agencies.
The authors declare no competing interests.
}

\vskip 0.2in
\bibliography{main}

\appendix
\crefalias{section}{appendix}
\crefname{appendix}{Appendix}{Appendices}
\Crefname{appendix}{Appendix}{Appendices}

\section{Dataset Documentation}
\label{sec:dataset_documentation}

This section provides documentation for the CIRCLED dataset following the recommendations of ``Datasheets for Datasets''~\citep{datasheets}.

\subsection{Access}
The dataset and code are publicly available:
\begin{itemize}
    \item \textbf{Dataset}: \url{https://huggingface.co/datasets/tk1441/CIRCLED}
    \item \textbf{Code}: \url{https://github.com/mti-lab/circled}
\end{itemize}

The dataset consists of multi-turn retrieval sessions in JSON format.
Each session consists of a sequence of (reference image ID, relative caption) pairs, paired with a ground truth image ID.
CIRCLED does not distribute the images themselves; it provides only image IDs and annotations (relative captions and session structures).
Users must download the original images from FashionIQ~\citep{fashioniq}, CIRR~\citep{cirr}, and CIRCO~\citep{circo} separately, following each dataset's terms of use.

\subsection{License}
The CIRCLED dataset annotations (relative captions and session structures) are released under the CC BY 4.0 license.
The underlying images are subject to their original licenses:
\begin{itemize}
    \item FashionIQ images: Subject to Amazon's terms of use
    \item CIRR/CIRCO images: Subject to original Flickr licenses (CC BY, CC BY-NC, etc.)
\end{itemize}

\subsection{Hosting and Maintenance Plan}
The dataset is hosted on Hugging Face Datasets, which provides long-term, reliable hosting with version control.
The code repository is maintained on GitHub.
We commit to maintaining the dataset for at least five years and will respond to issues and pull requests.
Any updates or corrections will be versioned and documented in the repository.

\subsection{Author Responsibility Statement}
The authors confirm that:
\begin{itemize}
    \item We bear all responsibility in case of violation of rights.
    \item The dataset annotations do not contain personally identifiable information.
    \item The relative captions were generated using GPT-4o-mini and reviewed through automated filtering to ensure quality and appropriateness.
\end{itemize}

\subsection{Image Provenance}
All images in CIRCLED originate from three existing, publicly available CIR datasets. These are FashionIQ~\citep{fashioniq} (fashion product images), CIRR~\citep{cirr} (real-life images from NLVR2), and CIRCO~\citep{circo} (COCO images). We do not collect any new images. CIRCLED contributes the multi-turn session structure and the LLM-generated relative captions on top of these sources. Per-image license information and, where required, generated stand-in images for figures are detailed in \cref{sec:image_licenses,sec:fashioniq_generation}, and each session records the source dataset and original image identifiers so that provenance can be traced back to the source datasets.

\subsection{Ethical Considerations}
Because CIRCLED reuses images from established public datasets and adds no new image collection, it introduces no new privacy concerns beyond those of the source datasets, and the annotations contain no personally identifiable information. The relative captions are produced by an LLM (GPT-4o-mini) and may reflect linguistic or cultural biases present in that model. We therefore fully release the construction prompts and filtering criteria so that such biases can be audited and improved by the community. As with any interactive retrieval resource, downstream systems trained or evaluated on CIRCLED (\eg e-commerce search) should be checked for fairness across user groups and product categories. Use of the dataset is subject to the terms of the underlying source datasets.

\section{Effect of History Integration Methods}
\label{sec:history_integration}

\begin{figure}[t]
  \centering
   \includegraphics[width=0.7\linewidth]{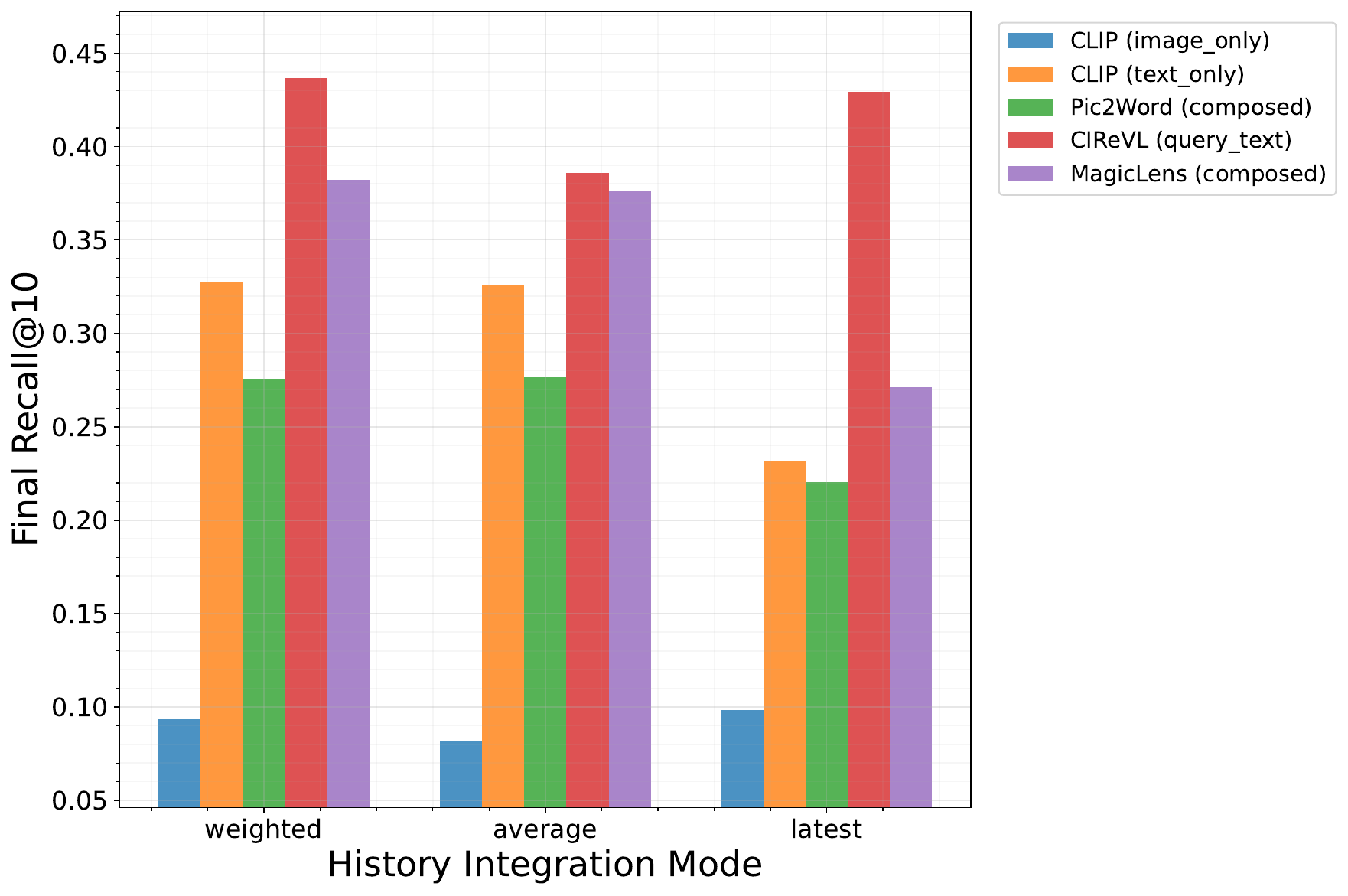}

   \caption{Example results for history-integration strategies.
  On the dataset extended from fashioniq\_dress\_val, we compare how different aggregation schemes affect each baseline's performance at each turn.}
   \label{historymode_comparison}
\end{figure}

We evaluate three ways of aggregating single-turn CIR features into a multi-turn method (Latest, Average, and Weighted) and report the results in terms of Final Recall@10.
As shown in \cref{historymode_comparison}, on the dataset extended from fashioniq\_dress\_val, the Weighted mode (which places larger weights on later turns) achieves the highest performance.
This aligns with the design of our dataset, where the query progressively approaches the target image in later turns, making that information particularly valuable.

Moreover, Weighted outperforms Latest (which uses only the final turn), indicating that leveraging the entire history yields more accurate retrieval than relying solely on the last turn.
An exception arises for image-only features, where Latest is slightly better, suggesting that the most recently selected reference image carries the most salient visual signal in that setting.

\section{Prompts}
\label{sec:prompts}
We list the prompts used in this work.
All prompts are fed into GPT-4o-mini.

\subsection{Caption Generation}
The prompts for generating image captions vary by dataset domain.

\noindent\textbf{FashionIQ Dress:}
\begin{lstlisting}
Describe this dress in 1-2 sentences. Focus on color, style, length, and key design features.
\end{lstlisting}

\noindent\textbf{FashionIQ Shirt:}
\begin{lstlisting}
Describe this shirt in 1-2 sentences. Focus on color, style, collar, sleeves, and key design features.
\end{lstlisting}

\noindent\textbf{FashionIQ Toptee:}
\begin{lstlisting}
Describe this top in 1-2 sentences. Focus on color, style, neckline, sleeves, and key design features.
\end{lstlisting}

\noindent\textbf{CIRR:}
\begin{lstlisting}
Describe this image in 1-2 sentences. Include main objects, people, setting, and notable features.
\end{lstlisting}

\noindent\textbf{CIRCO:}
\begin{lstlisting}
Describe this image in 1-2 sentences. Include main objects, composition, and notable features.
\end{lstlisting}

\subsection{Relative Caption Generation}
The prompt for generating relative captions includes history of previous suggestions to ensure diversity.
The retry mechanism is activated when the generated caption is too similar to existing ones.

\begin{lstlisting}
IMPORTANT -- Previous changes have already been suggested:
- "{relative caption1}"
- "{relative caption2}"
...
Your task is to identify a COMPLETELY DIFFERENT visual change.
Focus on aspects that have NOT been mentioned before.

[Only if retrying]
RETRY #{n}: The previous suggestion was too similar to existing ones.
Please provide a MORE DISTINCTIVE and DIFFERENT instruction.
Consider completely different visual aspects like:
- Different objects or people
- Different colors or lighting
- Different actions or poses
- Different background elements
- Different clothing or accessories

You will see two images.
**Image 1**: This is the REFERENCE image that needs to be modified.
**Image 2**: This is the TARGET image showing the desired result.

Write exactly ONE imperative instruction to transform Image 1 into Image 2.
Requirements:
1. Start with a verb
2. Be extremely specific about colors, positions, or actions
3. Avoid relative terms like "left" or "right"
4. Do not use quotes or explanatory text
5. Focus on a single, clear change
6. Must be different from previous suggestions
\end{lstlisting}

\subsection{Auxiliary Caption Generation}
The prompt for merging the reference image caption with the relative caption to create an auxiliary caption for retrieval.

\begin{lstlisting}
A user is performing image retrieval. The user provides a reference image
caption and a modification for the retrieved image to refine the search.
Generate a new query reflecting this modification. Only return the refined query.

Reference image caption: {reference_image_caption}
Modification: {relative_text}
\end{lstlisting}

\section{LLM-as-a-Judge Prompt for Session Evaluation}
\label{sec:llm_judge_prompt}

We use GPT-5-mini as an LLM-based judge to evaluate multi-turn retrieval sessions.
The system prompt is shown below.

\begin{lstlisting}
You are an expert annotator for multi-turn image retrieval dialogs.

You will see a complete retrieval session:
- A sequence of (CURRENT image, USER UTTERANCE) pairs showing the dialog progression
- A TARGET image (the final goal Z)

Your job is to evaluate the ENTIRE SESSION as a coherent dialog for image retrieval.

Rate the session on these 5 dimensions, each from 1 (very bad) to 5 (excellent):
1. SESSION NATURALNESS: Are the utterances consistently human-like throughout the session?
2. COHERENCE / CONSISTENCY: Is there logical flow without contradictions or abrupt changes?
3. GOAL-DIRECTED PROGRESS: Does each turn move closer to the target image Z?
4. REDUNDANCY (higher = better): Is there low repetition of information?
5. OVERALL: Holistic quality of the entire session
\end{lstlisting}

\section{Image Generation for FashionIQ Data}
\label{sec:fashioniq_generation}

For licensing reasons, in this paper we use generated images in the figures about FashionIQ dataset.
The images are generated by GPT-4o image generation function, using text prompts that describe the images in FashionIQ dataset.

\section{Scalability and Constraints}
\label{sec:scalability}

\Cref{tab:scalability} summarizes the computational cost and the main constraints of CIRCLED, supporting the discussion in the Broader Impact statement.

\begin{table}[t]
\centering
\caption{Scalability characteristics and constraints of CIRCLED.}
\label{tab:scalability}
\begin{tabular}{@{}lp{0.66\textwidth}@{}} \toprule
Aspect & Summary \\ \midrule
Construction cost & Bounded LLM (GPT-4o-mini) calls per session, namely one caption ($f_{\mathrm{caption}}$), one merge ($f_{\mathrm{merge}}$), and one difference per added turn ($f_{\mathrm{diff}}$), i.e., $O(L)$ calls for an $L$-turn session. \\
Query / evaluation cost & Training-free. Each turn is a single CLIP forward pass plus nearest-neighbor search over the subset database, with no per-dataset fine-tuning, and the history aggregation adds no learnable parameters. \\
Search-space scale & Retrieval scales with the per-subset database size (\cref{tab:turn_distribution_and_category}) using standard approximate nearest-neighbor indexing. \\
Constraints & Captions are English and LLM-generated, so they may carry linguistic bias. The evaluation trajectory is fixed and non-interactive. Some fashion source images are subject to link rot and the source datasets' terms of use. \\ \bottomrule
\end{tabular}
\end{table}

\section{Construction-Model Bias Analysis}
\label{sec:construction_bias}

A concern is that CIRCLED's rank progression might be specific to the construction baseline (CIReVL with a BLIP encoder), favoring its embedding geometry. We assess whether the progression holds under a retrieval setup that does \emph{not} use the construction encoder. For every session we compute, under several models, the $\varepsilon$-consistency satisfaction rate (fraction of sessions with $r_{l+1}\le r_l+\varepsilon$ at all turns, $\varepsilon=30$) and the median ground-truth rank per turn. All quantities are derived from the baseline evaluations already reported in \cref{sec:experiment results} and require no additional retrieval. Crucially, we include CIReVL with a CLIP encoder (the same construction algorithm with a \emph{different} encoder than the BLIP used for construction).

\Cref{tab:construction_bias} reports the result under a fixed CLIP encoder. CIReVL attains 92.6\%, clearly exceeding the non-CIReVL methods (80--83\%) and Image-only (63.8\%). Since CLIP is not the encoder used during construction, this shows that the rank progression is governed by the iterative query-update algorithm and holds under a different embedding space, rather than being tied to the construction encoder.

\begin{figure}[t]
\centering
\includegraphics[width=\linewidth]{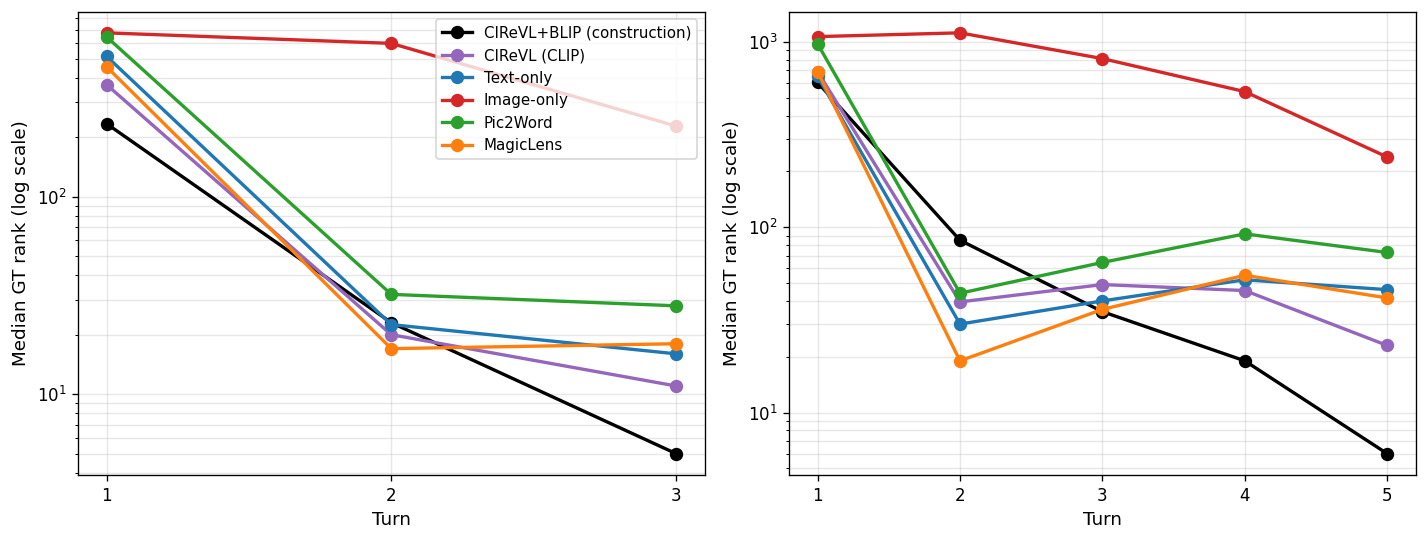}
\caption{Median ground-truth rank by turn under different retrieval models (log scale, pooled over subsets), for 3-turn (left, $N=5{,}424$, 24.0\% of data) and 5-turn (right, $N=772$, 3.4\%) sessions. All text-using models improve sharply at Turn~2. CIReVL (both encoders) continues to improve, while non-CIReVL methods maintain rank on the harder long sessions. Image-only shows little progression.}
\label{fig:construction_bias}
\end{figure}

\Cref{fig:construction_bias} shows the per-turn median rank. The Turn-1$\to$2 improvement is universal. Across all text-using models, the median rank drops from several hundred to single or low-double digits. Session length is determined by how many turns the construction baseline required, so longer sessions are intrinsically harder. The Turn-1 median rank generally grows with length (e.g., Text-only medians of 225, 514, 688, 650, and 951 for 2-, 3-, 4-, 5-, and 6-turn sessions). On these harder long sessions, non-CIReVL models improve at Turn~2 and then maintain the rank (e.g., for 5-turn sessions, Text-only goes $650\to30\to40\to52\to46$) rather than collapse, and over 80\% remain $\varepsilon$-consistent. Image-only retrieval shows little progression at any length, confirming that the textual modification drives the progression. We therefore position CIRCLED as a controlled benchmark whose progression structure holds across the retrieval methods we evaluate, not only the construction baseline.

\begin{table}[t]
\centering
\caption{$\varepsilon$-consistency satisfaction rate under a fixed CLIP encoder (\emph{not} the BLIP encoder used for construction), over all 22{,}608 CIRCLED sessions ($\varepsilon=30$). CIReVL clearly exceeds the non-CIReVL methods, showing the progression is driven by the iterative query-update algorithm and holds under a different encoder.}
\label{tab:construction_bias}
\begin{tabular}{@{}lc@{}} \toprule
Model (CLIP encoder) & $\varepsilon$-consistency \\ \midrule
CIReVL                & 92.6\% \\
Text-only             & 82.9\% \\
MagicLens             & 82.2\% \\
Pic2Word              & 80.5\% \\
Image-only            & 63.8\% \\ \bottomrule
\end{tabular}
\end{table}

\section{Human Evaluation of Session Quality}
\label{sec:human_eval}

To validate the LLM-as-a-judge, we conducted a human study on a 50-session subset of those it evaluated, using the same five dimensions and scoring rubric. The annotation was carried out by five researchers across two institutions (four male and one female), all independent of both the GPT-4o-mini used for generation and the GPT-5-mini used for judging. They had seen CIRCLED sessions during dataset construction, so we blinded the study to keep this familiarity from favoring CIRCLED. The 50 sessions were sampled with a fixed seed, balanced across the four sources (13/13/12/12), and shuffled, and each appeared under an anonymous identifier with the source method and the judge's ratings withheld from the annotators' kit. Matching the judge's zero-shot setting, annotators saw only the scoring rubric without worked examples, and they rated separately without discussing individual sessions. \Cref{tab:human_eval} reports the per-source means averaged over annotators. CIRCLED ranks first on all five dimensions. At the session level, the human ratings agree with the LLM judge (Spearman $\rho=0.65$ on Overall, over the 50 sessions). Notably, the annotators reproduce the same source ranking as GPT-5-mini exactly (Spearman $\rho=1.00$ over the four source means), with strong concordance among annotators (Kendall's $W=0.81$) and a significant Friedman test ($\chi^2=12.1$, $p<0.01$). Thus human annotators reproduce the judge's ranking at both the session and source levels, indicating the conclusions are not an artifact of self-preference between the generator and the judge.

\begin{table}[t]
\centering
\caption{Human evaluation of session quality (five annotators, 50 blind sessions, same rubric as the LLM judge). Mean rating per source ($1$--$5$, higher is better), best per column in bold. Human ratings agree with GPT-5-mini at the session level (Spearman $\rho=0.65$ on Overall) and reproduce the same source ranking exactly (Spearman $\rho=1.00$ over the four source means). Sources differ significantly (Friedman $p<0.01$).}
\label{tab:human_eval}
\begin{tabular}{@{}lccccc@{}} \toprule
Source & Natural. & Coher. & Goal Prog. & Non-redund. & Overall \\ \midrule
\textbf{CIRCLED (ours)}   & \textbf{4.25} & \textbf{4.15} & \textbf{4.40} & \textbf{4.45} & \textbf{4.22} \\
w/o quality filter        & 4.12 & 4.05 & 4.05 & 4.09 & 3.97 \\
Multi-turn FashionIQ      & 3.57 & 3.62 & 3.72 & 4.23 & 3.65 \\
Simple Concat             & 3.07 & 3.03 & 3.10 & 3.65 & 3.05 \\ \bottomrule
\end{tabular}
\end{table}

\begin{table}[t]
\centering
\caption{Statistical significance of CIRCLED's advantage over each comparison source (one-sided Mann-Whitney $U$ test on the 1{,}711 session ratings of \cref{tab:user_study}). We report $p$-values and rank-biserial effect sizes $r$ ($+$ favors CIRCLED). Here, n.s.\ denotes not significant.}
\label{tab:significance}
\begin{tabular}{@{}llcc@{}} \toprule
Comparison & Metric & $p$ & $r$ \\ \midrule
vs.\ w/o quality filter & Overall       & $1.6\times10^{-5}$  & $+0.14$ \\
                        & Coherence     & $1.0\times10^{-4}$  & $+0.13$ \\
                        & Goal Progress & $5.7\times10^{-9}$  & $+0.19$ \\
                        & Redundancy & $3.5\times10^{-6}$  & $+0.08$ \\
                        & Naturalness   & n.s.                & $-0.09$ \\ \midrule
vs.\ Multi-turn FashionIQ & Overall       & $1.5\times10^{-12}$ & $+0.26$ \\
                          & Coherence     & $9.3\times10^{-5}$  & $+0.14$ \\
                          & Goal Progress & $5.5\times10^{-17}$ & $+0.31$ \\
                          & Redundancy & $6.3\times10^{-9}$  & $+0.11$ \\
                          & Naturalness   & $9.1\times10^{-11}$ & $+0.24$ \\ \midrule
vs.\ Simple Concat & Overall       & $1.4\times10^{-58}$ & $+0.58$ \\
                   & Coherence     & $5.2\times10^{-44}$ & $+0.51$ \\
                   & Goal Progress & $9.0\times10^{-64}$ & $+0.61$ \\
                   & Redundancy & $1.4\times10^{-22}$ & $+0.23$ \\
                   & Naturalness   & $5.0\times10^{-19}$ & $+0.31$ \\ \bottomrule
\end{tabular}
\end{table}

\section{Sensitivity to the Filtering Thresholds}
\label{sec:sensitivity}

We analyze the dataset's sensitivity to the two filtering thresholds by re-running the full filtering pipeline while varying one threshold and fixing the other (the adopted $\varepsilon=30$, $\tau=0.8$ reproduces the 22{,}608 sessions of \cref{tab:turn_distribution_and_category}). As shown in \cref{tab:sensitivity}, the rank-margin threshold $\varepsilon$ has a negligible effect ($\pm0.6\%$ over $\varepsilon\in[10,100]$). The dataset is robust to its choice, consistent with the rank-margin filter removing only 562 sessions overall. In contrast, the diversity threshold $\tau$ governs dataset size, changing it $3.6\times$ over $\tau\in[0.7,0.9]$. It controls the trade-off between scale and per-turn novelty. Our choice ($\varepsilon=30$, $\tau=0.8$) lies at a balanced operating point.

\begin{table}[t]
\centering
\caption{Sensitivity of the final dataset size to the filtering thresholds, varying one while fixing the other. The adopted setting ($\varepsilon=30$, $\tau=0.8$) reproduces the 22{,}608 sessions of \cref{tab:turn_distribution_and_category}. $\varepsilon$ has little effect, while $\tau$ governs dataset size.}
\label{tab:sensitivity}
\begin{tabular}{@{}cc cc@{}} \toprule
\multicolumn{2}{c}{$\varepsilon$ (with $\tau=0.8$)} & \multicolumn{2}{c}{$\tau$ (with $\varepsilon=30$)} \\
\cmidrule(lr){1-2} \cmidrule(lr){3-4}
$\varepsilon$ & \#sessions & $\tau$ & \#sessions \\ \midrule
10  & 22{,}485 & 0.70 & 7{,}932 \\
20  & 22{,}561 & 0.75 & 14{,}066 \\
\textbf{30} & \textbf{22{,}608} & \textbf{0.80} & \textbf{22{,}608} \\
50  & 22{,}649 & 0.85 & 26{,}507 \\
100 & 22{,}677 & 0.90 & 28{,}380 \\ \bottomrule
\end{tabular}
\end{table}

\section{Choice of the Construction Generator (GPT-4o-mini)}
\label{sec:generator_choice}

\begin{figure}[t]
\centering
\begin{subfigure}{0.42\linewidth}
    \centering
    \includegraphics[width=0.7\linewidth]{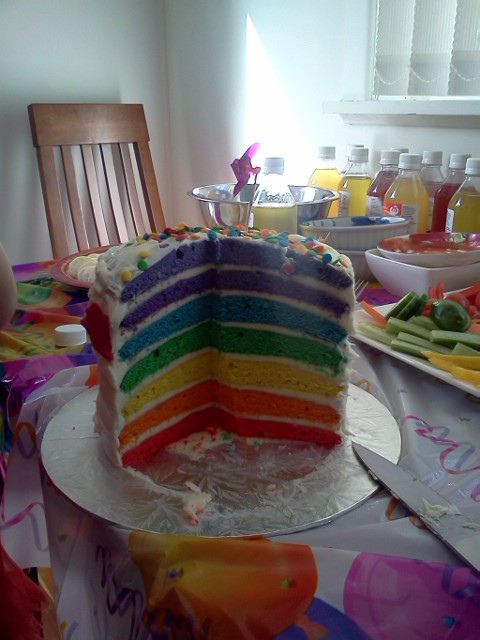}
    \caption{General (COCO \texttt{000000148864})}
    \label{fig:gen_example_coco}
\end{subfigure}
\hspace{0.04\linewidth}
\begin{subfigure}{0.42\linewidth}
    \centering
    \includegraphics[width=0.7\linewidth]{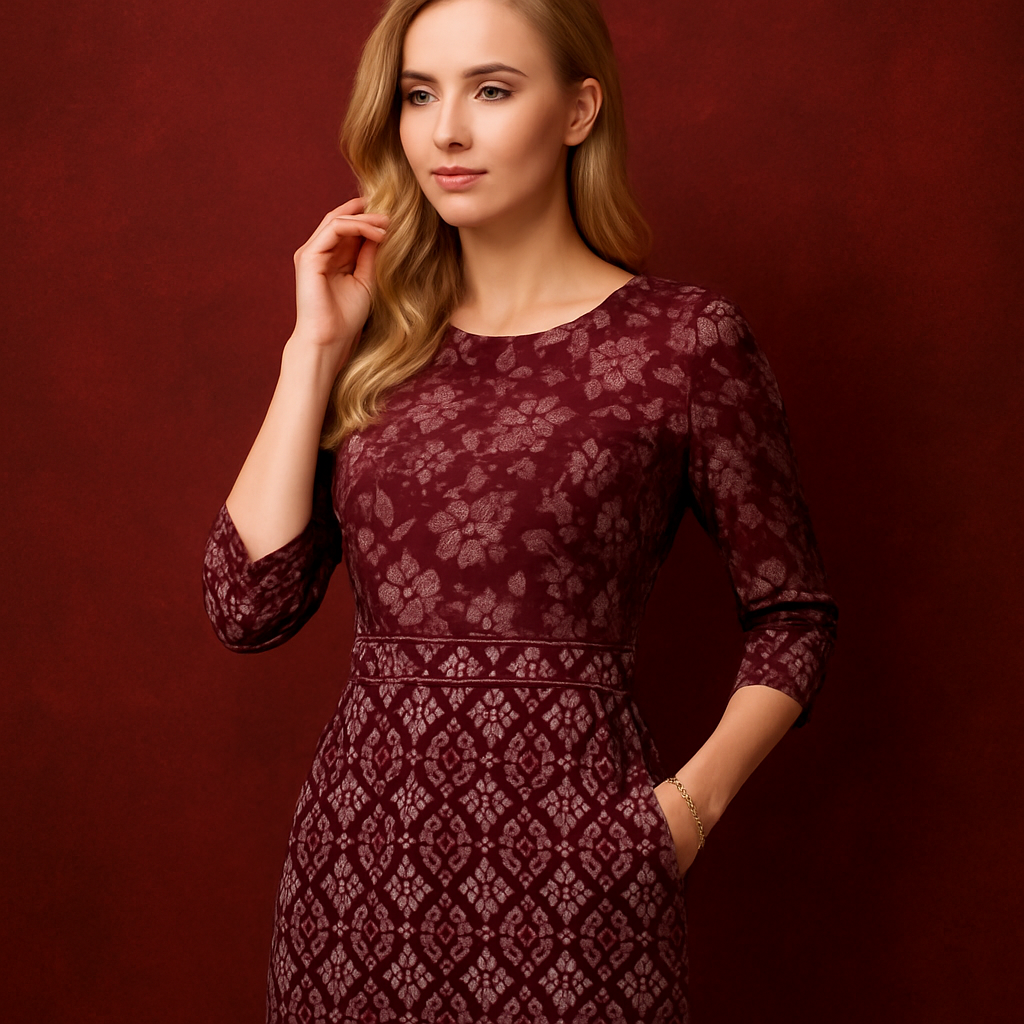}
    \caption{Fashion (FashionIQ \texttt{B00CBBV3ZC}, generated)}
    \label{fig:gen_example_fashion}
\end{subfigure}
\caption{Source images for the caption comparison in \cref{tab:caption_comparison}, both already used in this paper (\cref{fig:successful_examples}). The fashion image is the generated stand-in used in this paper (\cref{sec:fashioniq_generation}, with the original subject to the source dataset's terms). The COCO image is the original (CC BY-NC).}
\label{fig:caption_comparison_imgs}
\end{figure}

CIRCLED's construction uses GPT-4o-mini for the caption, merge, and difference generators ($f_{\mathrm{caption}}, f_{\mathrm{merge}}, f_{\mathrm{diff}}$). The LLM judge in \cref{sec:session_quality} is a different model, GPT-5-mini. To justify this choice we compare, on identical images, the captions produced by GPT-4o-mini and two open alternatives (BLIP2 and LLaVA). Verified against the source images, GPT-4o-mini accurately reports the fine-grained attributes relevant to composed retrieval in both the general and fashion domains (\cref{fig:caption_comparison_imgs}, \cref{tab:caption_comparison}). For the rainbow cake it gives the correct layer-color order and setting, and for the burgundy dress it correctly separates the floral bodice from the geometric skirt. By contrast, BLIP2 frequently hallucinates a different scene, and LLaVA produces short, subjective captions that, for fashion, often describe the model rather than the garment. In addition, only an instruction-following model can serve $f_{\mathrm{merge}}$/$f_{\mathrm{diff}}$, which require text integration and difference description rather than plain captioning. Combined with its low cost and wide availability, these properties make GPT-4o-mini a suitable construction generator.

\begin{table}[t]
\centering
\caption{Captions generated by GPT-4o-mini and two alternatives (BLIP2, LLaVA) on the images in \cref{fig:caption_comparison_imgs} (one general-domain, one fashion), already used in this paper. GPT-4o-mini accurately captures the fine-grained, retrieval-relevant attributes. BLIP2 hallucinates a different scene, and LLaVA gives short or subjective captions. Verified against the source images.}
\label{tab:caption_comparison}
\small
\begin{tabular}{@{}p{0.12\textwidth}p{0.80\textwidth}@{}} \toprule
\multicolumn{2}{@{}l}{\textbf{General domain, sliced rainbow cake} (COCO \texttt{000000148864}, \cref{fig:caption_comparison_imgs})} \\ \midrule
GPT-4o-mini & ``A vibrant rainbow cake with distinct layers of red, orange, yellow, green, blue, and purple, covered in white frosting and sprinkles, on a silver cake board with a festive table setting.'' \emph{(accurate)} \\
BLIP2 & ``the front of a train station with some people waiting for trains'' \emph{(hallucinated)} \\
LLaVA & ``A rainbow cake celebration awaits!'' \emph{(correct topic, no detail)} \\ \midrule
\multicolumn{2}{@{}l}{\textbf{Fashion domain, burgundy dress} (FashionIQ \texttt{B00CBBV3ZC}, \cref{fig:caption_comparison_imgs})} \\ \midrule
GPT-4o-mini & ``Burgundy dress with a floral pattern on the upper half and a geometric design on the skirt, with fitted three-quarter sleeves and a mid-thigh length.'' \emph{(accurate)} \\
BLIP2 & ``white dress with red and blue stripes, black stars and butterflies'' \emph{(hallucinated)} \\
LLaVA & ``woman posed against a red background, blonde hair, glamour'' \emph{(describes the model, not the garment)} \\ \bottomrule
\end{tabular}
\end{table}

\section{Word Clouds of Relative Captions}
\label{sec:wordclouds}

\begin{figure}[t]
    \centering
    \begin{subfigure}[b]{0.32\textwidth}
        \centering
        \includegraphics[width=\textwidth]{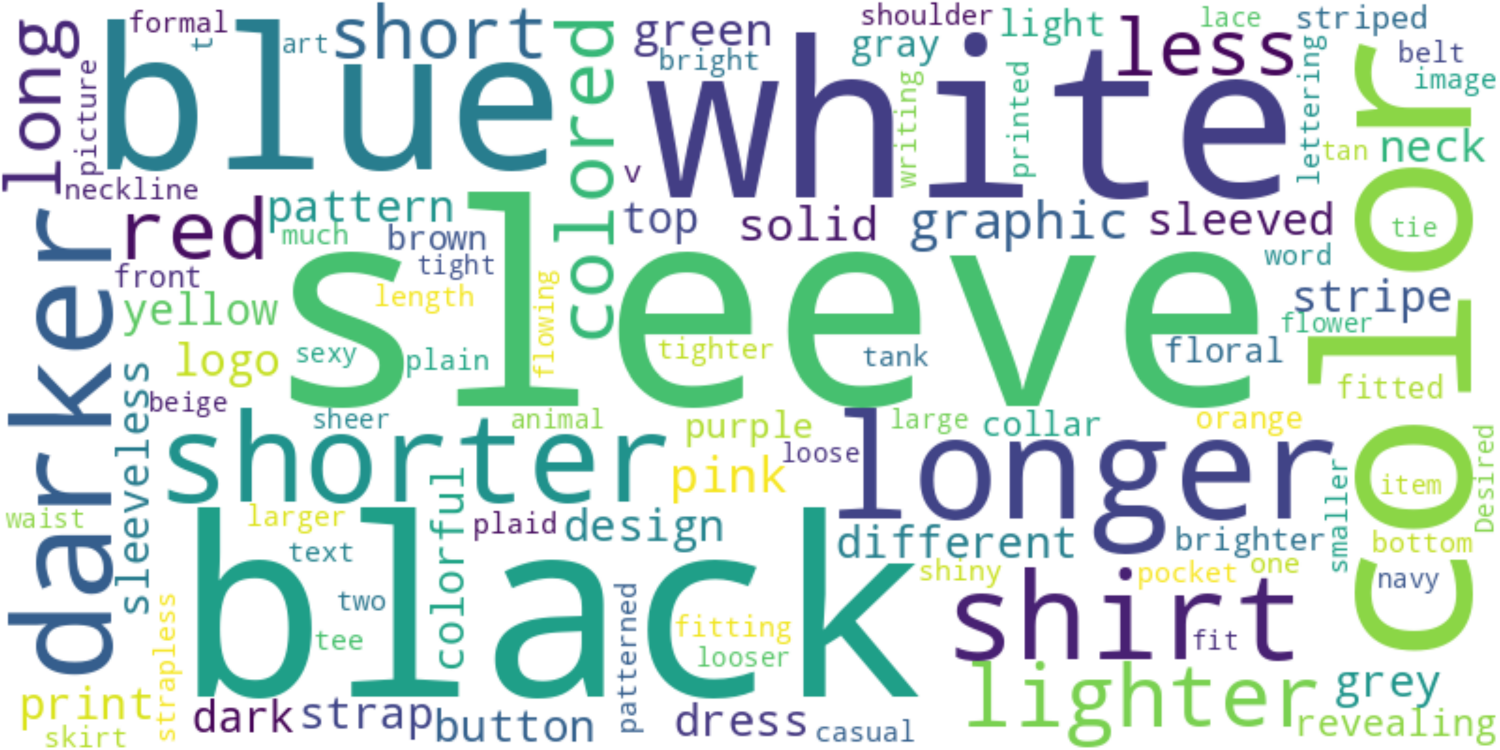}
        \caption{FashionIQ}
        \label{fig:wc_fashioniq}
    \end{subfigure}
    \hfill
    \begin{subfigure}[b]{0.32\textwidth}
        \centering
        \includegraphics[width=\textwidth]{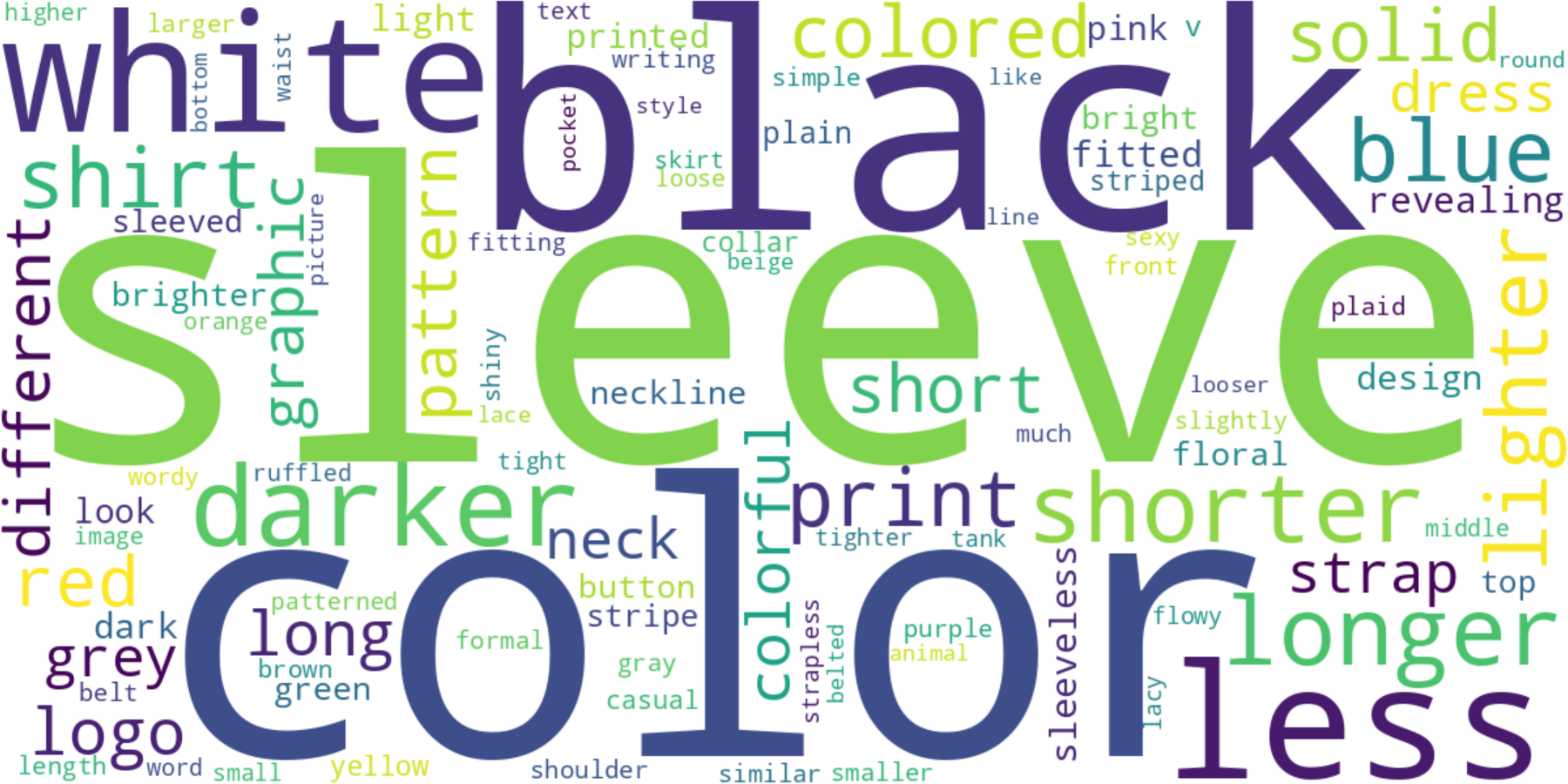}
        \caption{Multi-turn FashionIQ}
        \label{fig:wc_multiturn_fashioniq}
    \end{subfigure}
    \hfill
    \begin{subfigure}[b]{0.32\textwidth}
        \centering
        \includegraphics[width=\textwidth]{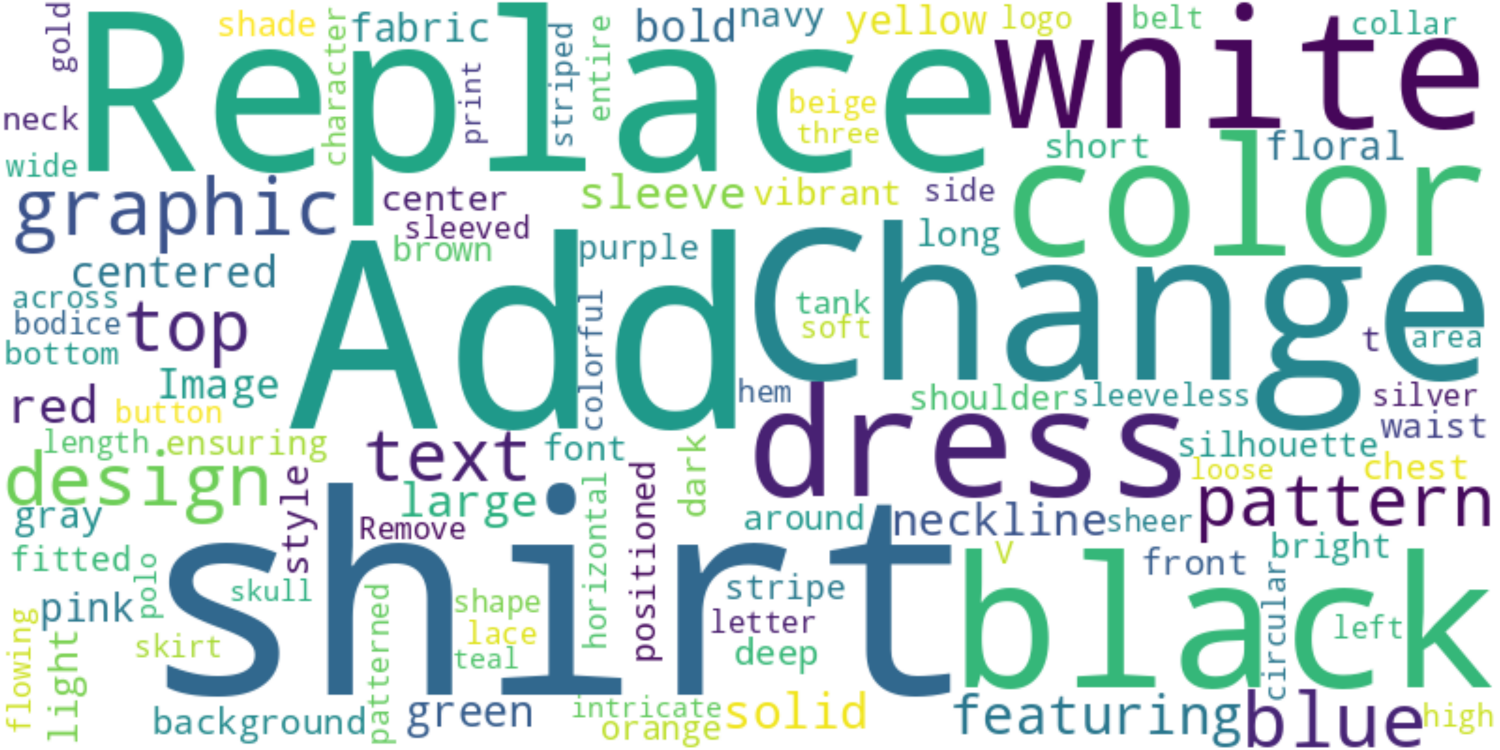}
        \caption{CIRCLED (Fashion)}
        \label{fig:wc_circled_fashion}
    \end{subfigure}

    \vspace{0.5em}

    \begin{subfigure}[b]{0.32\textwidth}
        \centering
        \includegraphics[width=\textwidth]{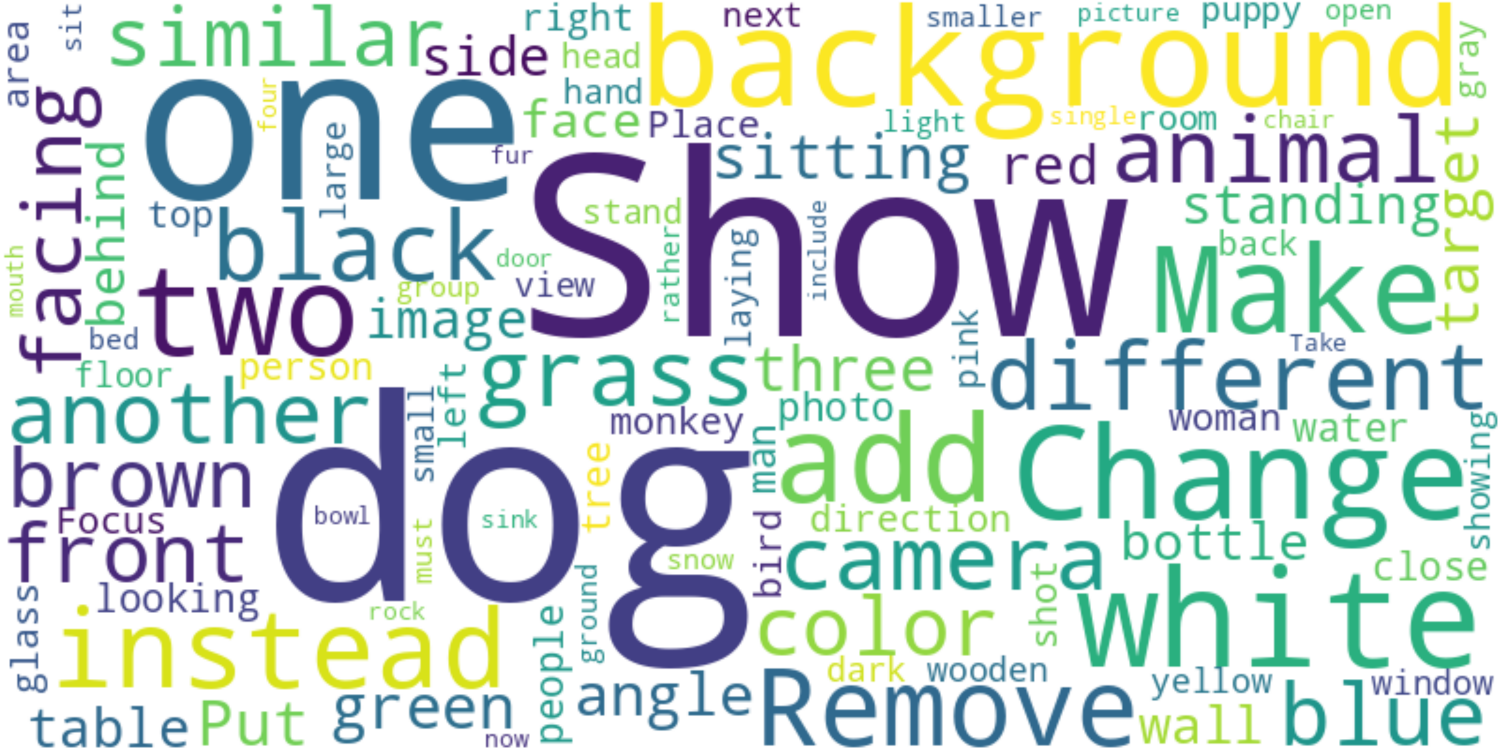}
        \caption{CIRR/CIRCO}
        \label{fig:wc_cirr_circo}
    \end{subfigure}
    \hspace{0.02\textwidth}
    \begin{subfigure}[b]{0.32\textwidth}
        \centering
        \includegraphics[width=\textwidth]{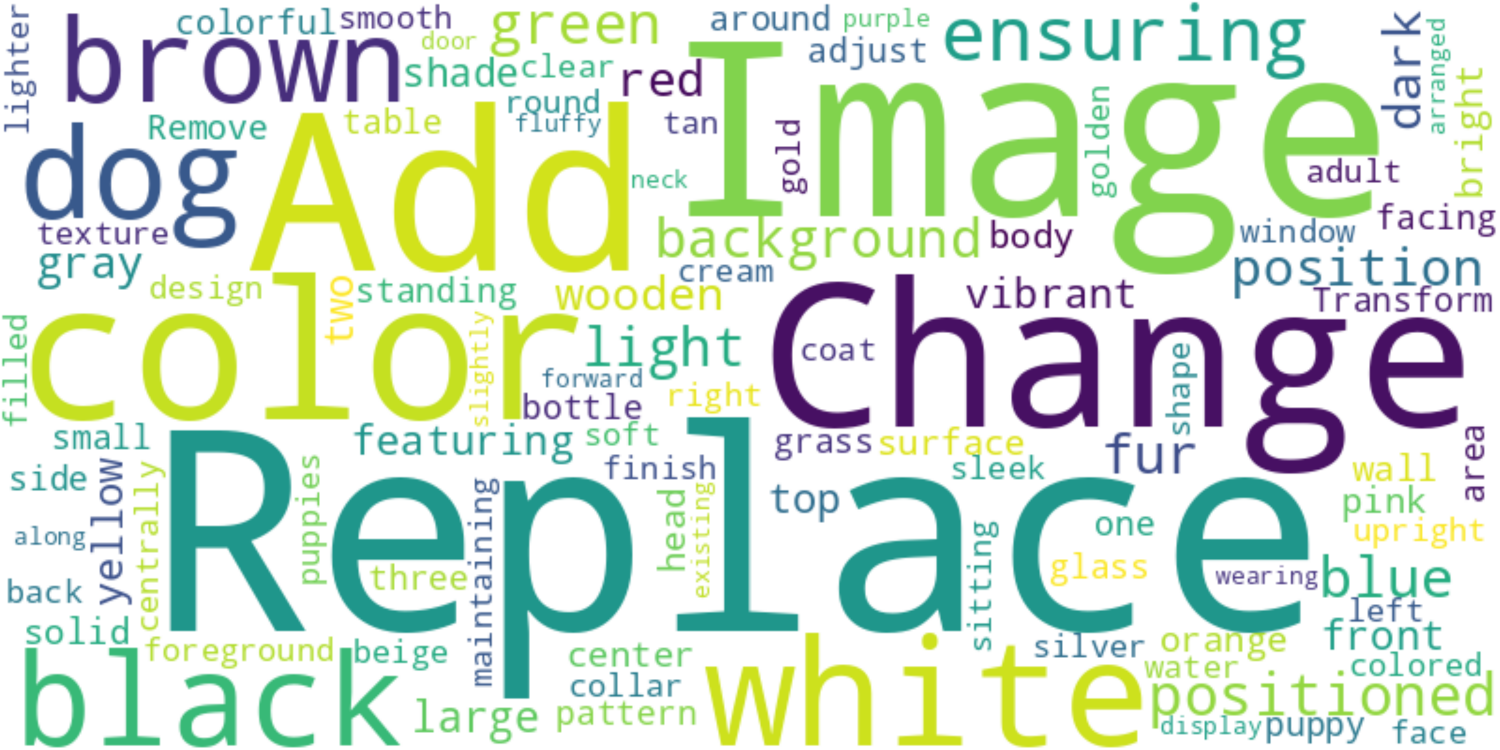}
        \caption{CIRCLED (General)}
        \label{fig:wc_circled_general}
    \end{subfigure}
    \caption{Word clouds of relative captions. Top row: Fashion domain (FashionIQ, Multi-turn FashionIQ, CIRCLED). Bottom row: General domain (CIRR/CIRCO, CIRCLED). FashionIQ and CIRR/CIRCO are single-turn CIR datasets, Multi-turn FashionIQ is a multi-turn dataset, and CIRCLED extends single-turn datasets to multi-turn.}
    \label{fig:wordclouds}
\end{figure}

\Cref{fig:wordclouds} complements the linguistic metrics of \cref{tab:linguistic_metrics} with a qualitative view of the vocabulary used in each dataset. The clouds are computed with the default stopword list of the Python WordCloud library.

\section{Image Licenses}
\label{sec:image_licenses}

The licenses of the images used in this paper are summarized in \cref{tab:image_ids,tab:image_licenses}.

\begin{table}
    \centering
    \resizebox{\textwidth}{!}{
    \begin{tabular}{@{}lll@{}} \toprule
        Figure & Images & Image ID \\ \midrule
        Fig.~\ref{fig:examples_multiturn_cir} & Turn1 reference image, search result & 000000190235, 000000037054\\
        Fig.~\ref{fig:examples_multiturn_cir} & Turn2 reference image, search result & 000000037054, 000000537111\\
        Fig.~\ref{fig:examples_multiturn_cir} & TurnN reference image, search result & 000000537111, 000000138205\\
        Fig.~\ref{fig:examples_multiturn_cir} & Ground truth & 000000138205\\[0.3em]

        Fig.~\ref{datasts_examples}~\subref{examples_of_our_datasets} & Reference image 1, 2, 3, ground truth & B004EHON6W, B004XJI4I4, B008D5AYOG, B0093K54X6\\[0.3em]

        Fig.~\ref{datasts_examples}~\subref{examples_of_multiturn_fashioniq} & Reference image 1, 2, 3, ground truth & B007XLHLHE, B008I5Q3CS, B007IXDC7U, B007IXDESW\\[0.3em]

        Fig.~\ref{flow_search_process}~\subref{flow_turn0} & Reference image, search results (from left)  & 000000112439, 000000518636, 000000324422, 000000264963\\
        Fig.~\ref{flow_search_process}~\subref{flow_turn1} & Non-selected images (from top), selected image  & 000000518636, 000000324422, 000000264963\\
        Fig.~\ref{flow_search_process}~\subref{flow_turn1} & Search results (from top)  & 000000460329, 000000346912, 000000247950, 000000264963\\[0.3em]

        Fig.~\ref{filter examples}~\subref{rank filter} & Reference image 1, 2, 3, 4, ground truth & 000000253571, 000000219002, 000000448025, 000000332048, 000000166818\\
        Fig.~\ref{filter examples}~\subref{relative caption filter} & Reference image 1, 2, 3, 4, ground truth & 000000447096, 000000271037, 000000230153, 000000195608, 000000478514\\[0.3em]

        Fig.~\ref{fig:successful_examples}~\subref{fig:example_circo} & Reference image 1, 2, ground truth & 000000179795, 000000318881, 000000148864\\[0.3em]

        Fig.~\ref{fig:successful_examples}~\subref{fig:example_fashioniq} & Reference image 1, 2, 3, ground truth & B008ZB39FY, B00CBBV3ZC, B0091S4Q9S, B008HQYFB4\\[0.3em]

        Fig.~\ref{fig:caption_comparison_imgs} & General image, fashion image (reused from \cref{fig:successful_examples}) & 000000148864, B00CBBV3ZC\\[0.3em]

        \bottomrule
    \end{tabular}
    }
        \caption{List of Image IDs used in Figures}
    \label{tab:image_ids}
\end{table}

\tiny
\begin{longtable}{@{}r>{\raggedright\arraybackslash}p{0.65\textwidth}l@{}}
    \caption{Image ID, URLs, and Licenses. If an image is noted ``generated image'' in its License column, it is generated by gpt-4o image generation function, using a text prompt that describes the image on the URL noted in its Image URL column.}
    \label{tab:image_licenses} \\
    \toprule
    Image ID & Image URL & License \\
    \midrule
    \endfirsthead
    \multicolumn{3}{c}{\tablename\ \thetable\ -- continued from previous page} \\
    \toprule
    Image ID & Image URL & License \\
    \midrule
    \endhead
    \midrule
    \multicolumn{3}{r}{Continued on next page} \\
    \endfoot
    \bottomrule
    \endlastfoot
        000000190235 & \url{http://farm7.staticflickr.com/6171/6207331658_b318513022_z.jpg} & CC BY 2.0 \\
        000000037054 & \url{http://farm7.staticflickr.com/6007/6198289264_029a6e88e2_z.jpg} & CC BY 2.0 \\
        000000537111 & \url{http://farm9.staticflickr.com/8064/8211635300_deba3583bf_z.jpg} & CC BY 2.0 \\
        000000138205 & \url{http://farm9.staticflickr.com/8175/8071171348_de3c9af840_z.jpg} & CC BY-NC-SA 2.0 \\
        000000112439 & \url{http://farm2.staticflickr.com/1137/962695681_0be4bcd0f8_z.jpg} & CC BY-NC-SA 2.0 \\
        000000518636 & \url{http://farm4.staticflickr.com/3460/4018979209_8dc1cf8ffd_z.jpg} & CC BY-NC-SA 2.0 \\
        000000324422 & \url{http://farm4.staticflickr.com/3244/2655366225_5b0754fb6e_z.jpg} & CC BY-NC-ND 2.0 \\
        000000264963 & \url{http://farm3.staticflickr.com/2334/2383999861_75d27c265e_z.jpg} & CC BY-NC-SA 2.0 \\
        000000460329 & \url{http://farm4.staticflickr.com/3182/3025936416_f1e5421a1b_z.jpg} & CC BY-NC-SA 2.0 \\
        000000346912 & \url{http://farm4.staticflickr.com/3323/3513955326_d9801f2b83_z.jpg} & CC BY 2.0 \\
        000000247950 & \url{http://farm5.staticflickr.com/4136/4743534549_c7a3612428_z.jpg} & CC BY-NC-SA 2.0 \\
        B004EHON6W & \url{http://ecx.images-amazon.com/images/I/41DXkZHTOjL.SX342.jpg} & generated image \\
        B0093K54X6 & \url{http://ecx.images-amazon.com/images/I/31r71PVOUEL.SX342.jpg} & generated image \\
        B004XJI4I4 & \url{http://ecx.images-amazon.com/images/I/31VgYvG6qbL.SX342.jpg} & generated image \\
        B008D5AYOG & \url{http://ecx.images-amazon.com/images/I/416DhASM95L.SX342.jpg} & generated image \\
        B007XLHLHE & \url{http://ecx.images-amazon.com/images/I/41LQcUsGHXL.SY445.jpg} & generated image \\
        B008I5Q3CS & \url{http://ecx.images-amazon.com/images/I/41rprim9AFL.SX342.jpg} & generated image \\
        B007IXDC7U & \url{http://ecx.images-amazon.com/images/I/41UxaR--QlL.SY445.jpg} & generated image \\
        B007IXDESW & \url{http://ecx.images-amazon.com/images/I/31BrgjaaIZL.SY445.jpg} & generated image \\
        000000253571 & \url{http://farm1.staticflickr.com/228/471397100_afd0fe517a_z.jpg} & CC BY-NC 2.0 \\
        000000219002 & \url{http://farm7.staticflickr.com/6038/6215343038_6bae3b45cb_z.jpg} & CC BY-NC-SA 2.0 \\
        000000448025 & \url{http://farm2.staticflickr.com/1019/1035634033_503438c7ea_z.jpg} & CC BY-NC-ND 2.0 \\
        000000332048 & \url{http://farm3.staticflickr.com/2320/2073978153_7d320747e4_z.jpg} & CC BY-NC 2.0 \\
        000000166818 & \url{http://farm3.staticflickr.com/2054/2247395963_7cb97cbf8d_z.jpg} & CC BY-NC-SA 2.0 \\
        000000447096 & \url{http://farm6.staticflickr.com/5001/5255997294_7df1664c69_z.jpg} & CC BY-NC-SA 2.0 \\
        000000271037 & \url{http://farm1.staticflickr.com/62/195329469_b95cf37cc0_z.jpg} & CC BY 2.0 \\
        000000230153 & \url{http://farm3.staticflickr.com/2405/2005336437_2e7b5da6db_z.jpg} & CC BY-NC-SA 2.0 \\
        000000195608 & \url{http://farm9.staticflickr.com/8302/7782408182_25d279cd27_z.jpg} & CC BY 2.0 \\
        000000478514 & \url{http://farm8.staticflickr.com/7252/7782273360_221f840574_z.jpg} & CC BY 2.0 \\
        000000179795 & \url{http://farm3.staticflickr.com/2737/4114913419_2736255889_z.jpg} & CC BY-NC-SA 2.0 \\
        000000148864 & \url{http://farm8.staticflickr.com/7080/7404128930_7b4c76a4e2_z.jpg} & CC BY-NC 2.0 \\
        000000318881 & \url{http://farm6.staticflickr.com/5528/10038641393_8f1ce796d0_z.jpg} & CC BY-NC-ND 2.0 \\
        B008ZB39FY & \url{http://ecx.images-amazon.com/images/I/41YqRey8zgL._SX342_.jpg} & generated image \\
        B008HQYFB4 & \url{http://ecx.images-amazon.com/images/I/31b\%2BheNk5nL._SX342_.jpg} & generated image \\
        B00CBBV3ZC & \url{http://g-ecx.images-amazon.com/images/G/01/x-locale/brands/lifestyle-assets/5370858011._CB335605943_SR150,160_.jpg} & generated image \\
        B0091S4Q9S & \url{http://ecx.images-amazon.com/images/I/31kZNo52IkL._SX342_.jpg} & generated image \\

\end{longtable}

\end{document}